\documentclass{article} % For LaTeX2e
\usepackage{iclr2026_conference,times}

\usepackage{amsmath,amsfonts,bm}

\def\eqref#1{equation~\ref{#1}}
\def\1{\bm{1}}

\DeclareMathAlphabet{\mathsfit}{\encodingdefault}{\sfdefault}{m}{sl}
\SetMathAlphabet{\mathsfit}{bold}{\encodingdefault}{\sfdefault}{bx}{n}

\usepackage{natbib}
\usepackage[utf8]{inputenc} % allow utf-8 input
\usepackage[T1]{fontenc}    % use 8-bit T1 fonts
\usepackage{hyperref}       % hyperlinks
\usepackage{url}            % simple URL typesetting
\usepackage{booktabs}       % professional-quality tables
\usepackage{amsfonts}       % blackboard math symbols
\usepackage{nicefrac}       % compact symbols for 1/2, etc.
\usepackage{microtype}      % microtypography
\usepackage{xcolor}         % colors
\usepackage{pifont}         % colors
\usepackage{booktabs, multirow, tabularx} 
\usepackage{colortbl} 
\usepackage{graphicx}
\usepackage{caption}
\usepackage{arydshln}
\definecolor{citecolor}{HTML}{0071bc}
\usepackage{amsmath}
\usepackage{cleveref}
\usepackage{xspace}  % 为了智能空格处理
\newcommand{\eg}{e.g.\xspace}
\usepackage{makecell}
\usepackage{threeparttable}
\usepackage{amssymb}

\usepackage{array}

\title{Massive Activations are the Key to Local Detail Synthesis in Diffusion Transformers}

\author{%
  Chaofan Gan$^{1,2}\quad$ Zicheng Zhao$^{1}\quad$ Yuanpeng Tu$^3\quad$ Xi Chen$^3\quad$ Ziran Qin$^1\quad$ \\ \textbf{Tieyuan Chen$^1\quad$} \textbf{Mehrtash Harandi$^2\quad$} \textbf{Weiyao Lin$^1\quad$}\\
  $^1$Shanghai Jiao Tong University, $^2$Monash University, $^3$The University of Hong Kong \\
  \href{https://ganchaofan0000.github.io/project_dg}{\texttt{\textcolor{magenta}{https://ganchaofan0000.github.io/DG}}}
}
\iclrfinalcopy % Uncomment for camera-ready version, but NOT for submission.
\begin{document}

\maketitle
\begin{figure}[htp]
  \vspace{-10mm}
  \centering
  \includegraphics[width =\linewidth]{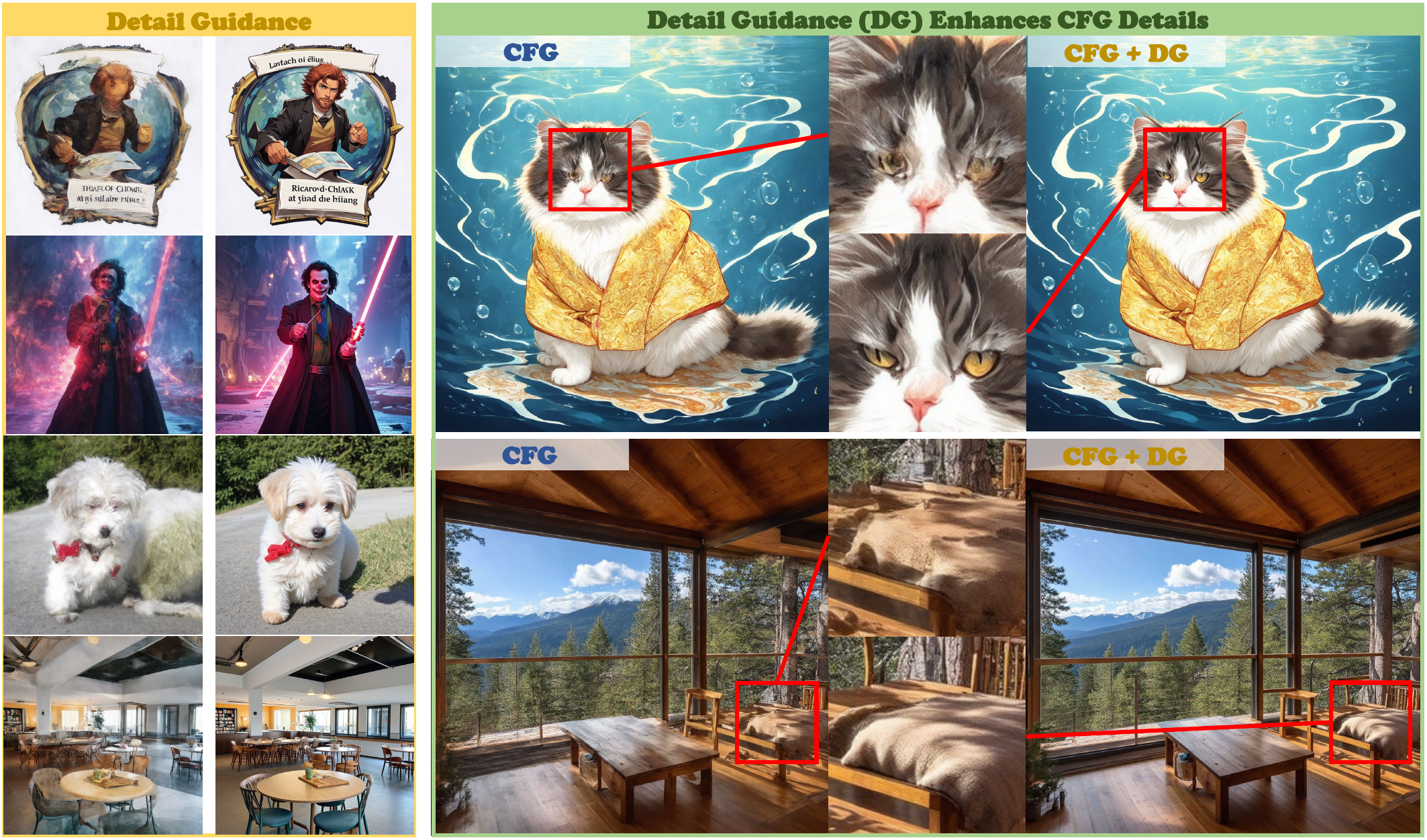}
  \vspace{-6mm}
  \caption{\textbf{Visual results of our Detail Guidance (DG).}  
    Left: DG explicitly enhances fine-grained visual details, yielding high-quality outputs.  
    Right: DG integrates seamlessly with Classifier-Free Guidance (CFG), allowing for further refinement of details.
    }
    \vspace{-1mm}
  \label{figure:f1}
  % \vspace{-10mm}
\end{figure}

\begin{abstract}
Diffusion Transformers (DiTs) have recently emerged as a powerful backbone for visual generation. 
Recent observations reveal \emph{Massive Activations} (MAs) in their internal feature maps, yet their function remains poorly understood. 
In this work, we systematically investigate these activations to elucidate their role in visual generation. 
We found that these massive activations occur across all spatial tokens, and their distribution is modulated by the input timestep embeddings. Importantly, our investigations further demonstrate that these massive activations play a key role in local detail synthesis, while having minimal impact on the overall semantic content of output.
Building on these insights, we propose \textbf{D}etail \textbf{G}uidance (\textbf{DG}), a MAs-driven, training-free self-guidance strategy to explicitly enhance local detail fidelity for DiTs. Specifically, DG constructs a degraded ``detail-deficient'' model by disrupting MAs and leverages it to guide the original network toward higher-quality detail synthesis. Our DG can seamlessly integrate with Classifier-Free Guidance (CFG), enabling further refinements of fine-grained details.
Extensive experiments demonstrate that our DG consistently improves fine-grained detail quality across various pre-trained DiTs (\eg, SD3, SD3.5, and Flux).

\end{abstract}

\section{Introduction}
Diffusion models~\citep{rombach2022high,saharia2022photorealistic} have recently achieved remarkable success across a wide range of generative tasks.
Among various architectures, the Transformer~\citep{vaswani2017attention} has emerged as a powerful and versatile backbone for diffusion models~\citep{peebles2023scalable}, thanks to its flexibility and scalability.
With the increasing availability of large-scale data and computational resources, many large Diffusion Transformers (DiTs)~\citep{peebles2023scalable,esser2024scaling} have recently emerged, achieving state-of-the-art performance in both image and video synthesis~\citep{yang2024cogvideox,hong2022cogvideo,wan2025}.

Along with the rapid progress of DiTs, recent studies~\citep{sun2024massive, darcet2023vision, gan2025unleashing} have uncovered an interesting phenomenon known as \textit{Massive Activations} (MAs) in these Transformer-based models, where rare hidden activations exhibit unusually large magnitudes. Specifically, ~\citep{sun2024massive, xiao2024efficient} identifies the massive activations in Large Language Models (LLMs) and demonstrates that they are essential for long-context learning. Similar activation patterns are observed in Vision Transformers (ViTs), where they are utilized to process global semantic information~\citep{darcet2023vision}. More recently, several works~\citep{gan2025unleashing, fang2025tinyfusion} have reported the presence of massive activations in DiTs. However, their functional role within the visual generation process of DiTs remains largely unexplored.

% \MH{please double check the refs. some of the arxiv papers are actually published such as Darcet-2023 and Gan-2025 :-)}

\begin{figure}[tp]
  \centering
  \includegraphics[width =\linewidth]{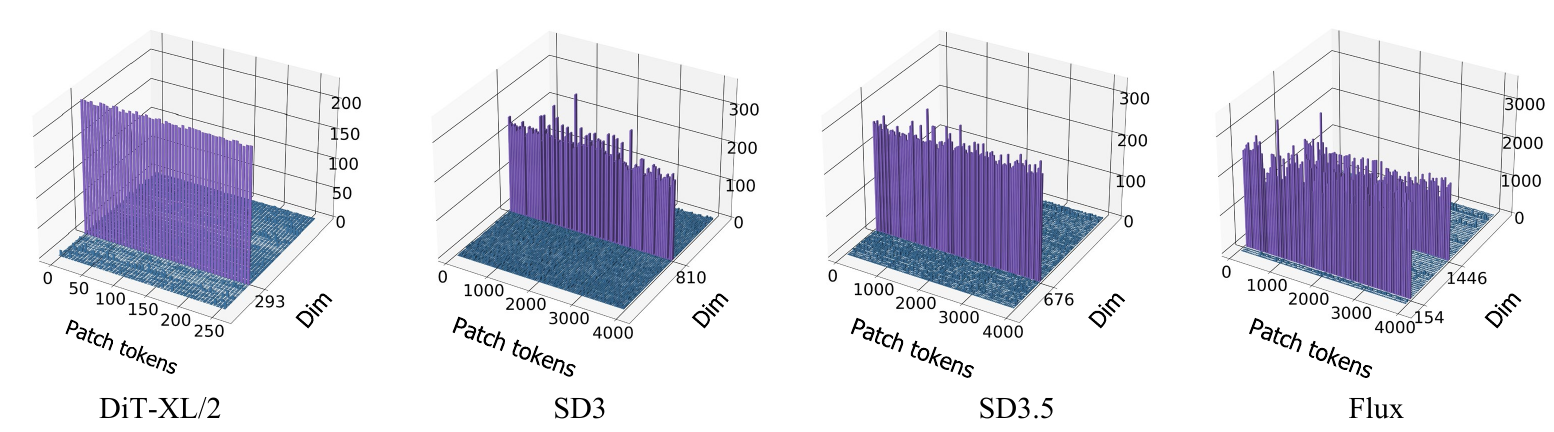}
  \caption{\textbf{Massive Activations in DiTs.}  
    The activation magnitudes of internal hidden states. We present the average magnitudes over 1{,}000 text prompts.
    Massive activations are consistently concentrated in a few fixed dimensions across all image patch tokens.
}
  \vspace{-8mm}
  \label{figure:f2}
\end{figure}

In this paper, we aim to gain a deeper understanding of the role massive activations play in the visual generation tasks. We first conduct systematic investigations to study the characteristics of massive activations. 
Our investigations reveal that massive activations appear in a few fixed dimensions across all image tokens, which are text-independent (~\Cref{figure:f2,figure:f4}). In addition, we demonstrate that these activations are closely associated with the input timestep embeddings, where the timestep encoding can directly shape its distribution (\Cref{figure:f4}).

Furthermore, we perform activation intervention by disrupting the internal massive activations to directly investigate their impact on DiT generation. Our analysis shows that, when disrupting the massive activations, the visual output preserves consistent semantic content with the original images (\Cref{figure:f5}). These results suggest that the massive activations exert minimal influence on the semantics of the generation process. On the other hand, we found that the local details of visual output are significantly degraded when massive activations are disrupted, suggesting their crucial role in local detail synthesis (\Cref{figure:f5}). We propose the following interpretation to these findings: \textit{DiT assigns massive activations to all spatial tokens to drive fine-grained local detail synthesis, while timestep embeddings modulate these activations to adaptively control the detail synthesis process throughout generation.}

% To probe the role of these token-wise massive activations in DiT visual generation, we perform a targeted intervention: we zero out the massive-activation dimensions at the first block and propagate the modified hidden states through the remaining network. The resulting samples exhibit pronounced texture loss and blurred fine details, indicating that massive activations directly support detail reconstruction. Further analysis of self-attention shows that these token-wise massive activations act as an implicit attention regularizer, increasing the effective attention mass received by each token, encouraging broader token participation during optimization, and ultimately improving the convergence of spatial details.

Motivated by these insights, we introduce \textbf{D}etail \textbf{G}uidance (\textbf{DG}), a MAs-driven, training-free self-guidance strategy for detail enhancement in DiT generation. Specifically, we construct a degraded ``detail-deficient'' network by disrupting the massive activations, and then leverage it to explicitly guide the original model toward generating higher-quality details. Our approach can be seamlessly integrated with classifier-free guidance (CFG), thereby achieving further detail refinement for CFG. Our main contributions can be summarized as follows.

\begin{itemize}
\item
We provide a comprehensive study of massive activations in DiTs, demonstrating that these activations are crucial for fine-grained local detail synthesis while exerting minimal influence on the overall semantic
content.
\item 
We trace the massive activations to the influence of the input timestep embeddings, revealing that the input timestep encoding can directly shape
their distribution.

\item
We introduce Detail Guidance (DG), a MA-driven, training-free self-guidance strategy to explicitly enhance local detail synthesis in DiTs.
DG integrates seamlessly with Classifier-Free Guidance (CFG), leading to improved local detail synthesis.
\end{itemize}

\section{Related Work}

% \MH{Chaofan, make sure your text is different enough from your previous work especially in sections such as related work or dataset description and etc. This is to avoid nasty reviewers complaining about similarity in text (I have seen this and very hard to rebut as you cannot say I was the author of the article you say my draft is similar to!}

\subsection{Diffusion Model}
Diffusion models~\citep{rombach2022high,dhariwal2021diffusion,saharia2022photorealistic} have become a dominant paradigm for high-quality visual synthesis. 
Early approaches primarily relied on U-Net architectures~\citep{rombach2022high} to model the denoising process. 
Recently, the field has shifted toward Transformer-based backbones~\citep{vaswani2017attention}. 
Among these advances, Diffusion Transformers (DiTs)~\citep{peebles2023scalable} have rapidly established themselves as a powerful backbone for visual generation. 
Due to the strong scalability and flexibility of transformer architecture, a new wave of large-scale DiTs\citep{esser2024scaling,Flux} (e.g., SD3, Flux) has emerged, achieving superior performance in various visual generation tasks~\citep{yang2024cogvideox,wan2025}.

\subsection{Massive Activations}
\noindent\textbf{Massive activations in LLMs.}
Recent studies~\citep{sun2024massive, zhao2023unveiling, xu2025slmrec} have identified the presence of massive activations in large language models (LLMs). These activations typically emerge at fixed dimensions of low-information tokens, such as starting or delimiter tokens. Importantly, some works~\citep{xiao2024efficient, jin2025massive} have shown that massive activations contribute positively to contextual knowledge modeling, enabling LLMs to capture long-range dependencies more effectively. In addition, \citep{jin2025massive} traced the emergence of concentrated massive values into rotary position embeddings (RoPE).

\noindent\textbf{Massive activations in ViTs.}
Similar activation patterns have also been observed in Vision Transformers (ViTs)~\citep{darcet2023vision, yang2024denoising, sun2024massive}. In ViTs, massive activations frequently arise in redundant background tokens and have been associated with encoding global semantic information. Moreover, \citep{yang2024denoising} traced the emergence of these activations to the influence of the input positional embeddings.

\noindent\textbf{Massive activations in DiTs.}
Several studies on the acceleration of Diffusion Transformers (DiTs)~\citep{liu2024hq,fang2025tinyfusion,zhao2024vidit} have highlighted the presence of outlier activations, whose extreme values pose a significant challenge for model quantization and distillation.
More recently, DiTF~\citep{gan2025unleashing} found that massive activations occur at fixed dimensions across all spatial tokens
when employing DiTs as feature extractors, and showed that these activations substantially influence the discriminative quality of extracted features.
However, the function of these massive activations in visual generation remains largely unexplored.
% %
% \MH{you should dexribe DiTF at an idea level and not about ADaLN. Say what Gan \etal observe and what they used it for (similar to the way you described it for LLMs)}

\subsection{Sampling guidance for diffusion models.} 
Classifier-free guidance (CFG)~\citep{ho2022classifier} has become the standard guidance mechanism for diffusion sampling. 
It extrapolates between the conditional and unconditional branches to amplify conditioning signals, thereby enhancing controllability and improving semantic alignment. Recently, auto-guidance~\citep{karras2024guiding} introduced a self-guidance signal by guiding the base model with a deliberately degraded “bad” version (e.g., reduced capacity or under-trained checkpoints), steering sampling toward higher-quality outputs. Other approaches~\citep{ahn2024self,hong2023improving,hyung2025spatiotemporal} construct degraded predictions by perturbing internal mechanisms such as modifying attention maps or skipping blocks to bias the sampler toward a better image distribution.

\section{Preliminaries}
\label{preliminary}

\noindent\textbf{Diffusion models.}
Diffusion models~\citep{ho2020denoising, karras2022elucidating} generate data by progressively denoising Gaussian noise, starting from $z_T \sim \mathcal{N}(0, I)$. Given a clean sample $x \sim p_{\text{data}}(x)$, the forward diffusion process can be expressed as $z_t = x + \sigma(t)\,\epsilon,$ where $\sigma(t)$ denotes the noise schedule and $\epsilon \sim \mathcal{N}(0, I)$. To learn the reverse process, a denoising network $D_{\theta}(z_t, t, c)$ is trained to predict the injected noise at each step, 
% The training objective is formulated as
% \begin{equation}
% \min_{\theta}\, \mathbb{E}_{x, \epsilon, t}\left[\left\|\epsilon - D_\theta(z_t, t, c)\right\|^2\right],
% \end{equation}
where $c$ is the conditioning signals (e.g., class labels or text prompts). 
% In addition, the denoising network also allows for unconditional generation by training network $D_{\theta}(z_t, t)$.

\noindent\textbf{Classifier-Free Guidance.}
Classifier-Free Guidance (CFG)~\citep{ho2022classifier} enhances diffusion model quality by jointly training the denoising network in conditional $D_{\theta}(z_t,t,c)$
and unconditional $D_{\theta}(z_t,t)$ modes.
At sampling time, it combines the two predictions to amplify the conditioning signal:
\begin{equation}
\hat{D}_{\theta}\left(z_t, t, c\right)=D_{\theta}\left(z_t, t\right)+w\left(D_{\theta}\left(z_t, t, c\right)-D_{\theta}\left(z_t, t\right)\right)
\end{equation}
where $w$ is the guidance scale. By extrapolating their difference, CFG strengthens semantic alignment and improves generation fidelity, but can sometimes lead to insufficient synthesis of fine-grained local details~\citep{sadat2024eliminating,chung2024cfg++}.

\noindent\textbf{DiT architecture.}
We provide the architecture of Diffusion Transformer (DiT) following~\citep{peebles2023scalable}. For clarity, we omit the VAE component and focus on the latent diffusion transformer, denoted as $D_{\theta}=\{D_k\}_{k=1}^{N}$, where $k$ indexes the block and $N$ is the total number of blocks. Given noised latent $z_t \in \mathbb{R}^{C \times H \times W}$, the DiT block $D_k$ forward the internal hidden state $z_t^k$ through a residual connection~\citep{he2016deep} to the next block, formulated as
\begin{equation}
z_{t}^{k+1} = z_{t}^{k}+\alpha_k D_k(z_{t}^{k}, t)
\label{block}
\end{equation}
where $\alpha_k \in \mathbb{R}^{C}$ denotes the dimension-wise residual scaling factor derived from the AdaLN layer~\citep{perez2018visual}. More architecture details can be found in~\Cref{sup:dit}.

% \MH{you need to define what $\mathcal{D}$ is (is it $D$ or something else, also is it describing the functionality applied or what). Also, it would be better to denote functions by lower-case alphabets such as f,h,g,...}

% \noindent\textbf{AdaLN in DiT.}
% Specifically, the AdaLN layer encodes the timestep embedding $t$ and additional conditioning signals $c$ into channel-wise modulation parameters through MLP networks:
% \begin{gather}
% \gamma_k, \beta_k, \alpha_k = \operatorname{MLP}_k(t, c)\\
% \hat{z}_{t}^{k}=\operatorname{AdaLN}(z_{t}^{k}; \gamma_k, \beta_k)
% \label{regress}
% \end{gather}
% where $\gamma_k$ and $\beta_k$ are applied to adaptively normalize the input latent $z_{t}^{k}$, while $\alpha_k$ scales the $k$-th residual connection.

\begin{figure}[tp]
  \centering
  \includegraphics[width =\linewidth]{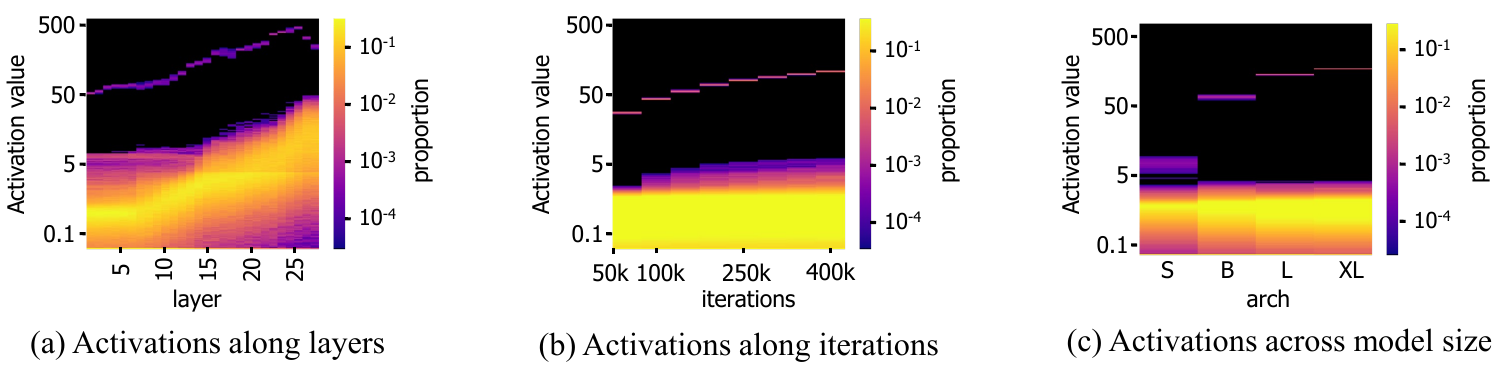}
  \caption{\textbf{Illustration of several properties of massive activations in DiT-XL.} (a) Activation distribution of the hidden states along DiT layers (b) Activation distribution of the hidden states along training iterations (c) Activation distribution of the hidden states across different model sizes. Massive activations occur throughout all layers and persist across different model sizes.}
  \vspace{-2mm}
  \label{figure:f3}
\end{figure}

\section{Massive Activations in Diffusion Transformers}
% \label{main:}
As shown in~\Cref{figure:f2}, the hidden states of various DiTs consistently exhibit a prominent phenomenon: \emph{Massive Activations} (MAs). This observation suggests that massive activations must play a crucial role in the visual generation process of DiTs. In this section, we conduct an in-depth investigation to understand \textit{why} and \textit{where} these massive activations emerge, and analyze their \textit{role} in the visual generation process of DiTs.

\subsection{Characteristics of Massive Activations}
\label{main:characteristics}
\paragraph{Massive activations occur throughout all layers across different model sizes.}
\vspace{-7pt}
We first investigate \textit{where} massive activations emerge. As shown in~\Cref{figure:f3}, we observed that massive activations exist in all internal DiT layers(\Cref{figure:f3}(a)). They emerge early during training (before 50k training iterations in \Cref{figure:f3}(b)), underscoring their crucial role in the internal computations of DiTs. Moreover, massive activations persist across models of different scales (\Cref{figure:f3}(c)). We present the layer properties of SD3, SD3.5, and Flux in~\Cref{sup:layer}, which further confirm their presence throughout all internal blocks.

\paragraph{Massive activations appear in fixed dimensions across all patch tokens.}
\vspace{-7pt}
% Unlike the massive activations in LLMs or ViTs~\citep{sun2024massive,darcet2023vision}, which typically arise in specific tokens such as start or delimiter tokens, massive activations in DiTs occur across all image tokens. To locate these activations in both class-to-image and text-to-image generation, 
Then, we analyze the spatial distribution of massive activations, as illustrated in~\Cref{figure:f2}. The results reveal that massive activations consistently appear at a fixed feature dimension (e.g., dimension 810 for SD3) across all spatial tokens. DiTF~\citep{gan2025unleashing} has also characterized similar properties.

% \paragraph{Timestep embeddings shape the distribution of massive activations.}
% \vspace{-5pt}
To delve into the massive activations in hidden states, we first examine the computation of hidden states $z_t^k \in \mathbb{R}^{C \times H \times W}$:
\begin{equation}
z_{t}^{k+1} = z_{t}^{k}+\alpha_k D_k(z_{t}^{k}, t), \alpha_k = \operatorname{MLP}_k(t, c)
\label{computation}
\end{equation}
where hidden states are computed via a residual connection, and $\alpha_k \in \mathbb{R}^{C}$ denotes the dimension-wise scaling factor parameters regressed by the AdaLN layer with an MLP network (see~\Cref{sup:dit} for details). As shown in~\Cref{figure:f4}(a), we compare the activation distributions of hidden states $z_t^k$ and the scaling factor $\alpha_k$ across dimensions. It can be observed that a prominent peak of $\alpha_k$ at dimension 810 leads to a corresponding concentration of massive activations (\Cref{figure:f4}(a)), indicating that the scaling factor $\alpha_k$ governs the dimension pattern of massive activations.

As the scaling factor $\alpha_k$ is produced by the AdaLN layer conditioned on the input timestep embedding $t$ and text embedding $c$ (see~\Cref{computation}),  
we further examine how $t$ and $c$ respectively influence MAs (\Cref{figure:f4}(b)).  
This analysis leads to two key observations:

\begin{figure}[tp]
  \centering
  \includegraphics[width =\linewidth]{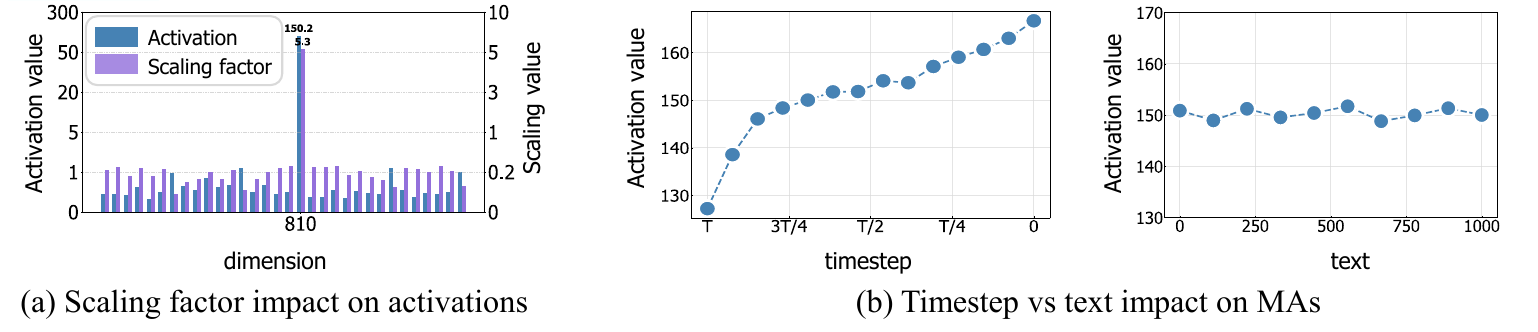}
  \caption{\textbf{Impact of the input timestep and text on Massive Activations (MAs) in SD3.}
(a) Comparison of the distributions of hidden-state $z_t^k$ activations and their corresponding residual scaling factor $\alpha_k$.
(b) Respective impact of input timestep and text embeddings on the magnitude distribution of MAs, where we compare the MAs of 1000 different text inputs. The massive activations are governed by the residual scaling factor; their magnitude is primarily shaped by the input timestep embedding $t$, while text embeddings $c$ have negligible effect.}
  \vspace{-5mm}
  \label{figure:f4}
\end{figure}

\paragraph{Text embeddings have minimal impact on massive activations.}
\vspace{-7pt}
As shown in~\Cref{figure:f4}(b), we compare the massive activation value across 1{,}000 different text prompts. We observe that these activations remain nearly constant (around 150) regardless of the input text embeddings,
indicating that the text embeddings have negligible influence on the magnitude of the massive activations.
\paragraph{Timestep embeddings shape the massive activations.}
\vspace{-7pt}
In contrast, we find that the timestep $t$ plays a dominant role for massive activation: the magnitude of massive activations increases steadily as $t$ decreases from $T$ to $0$. We also get similar observations for SD3-5 and Flux (see~\Cref{sup:timestep}). These results suggest that massive activations in DiTs are mainly modulated by the timestep embeddings.

\subsection{Massive Activations for Local Detail Synthesis}
\begin{figure}[tp]
  \centering
  \includegraphics[width =\linewidth]{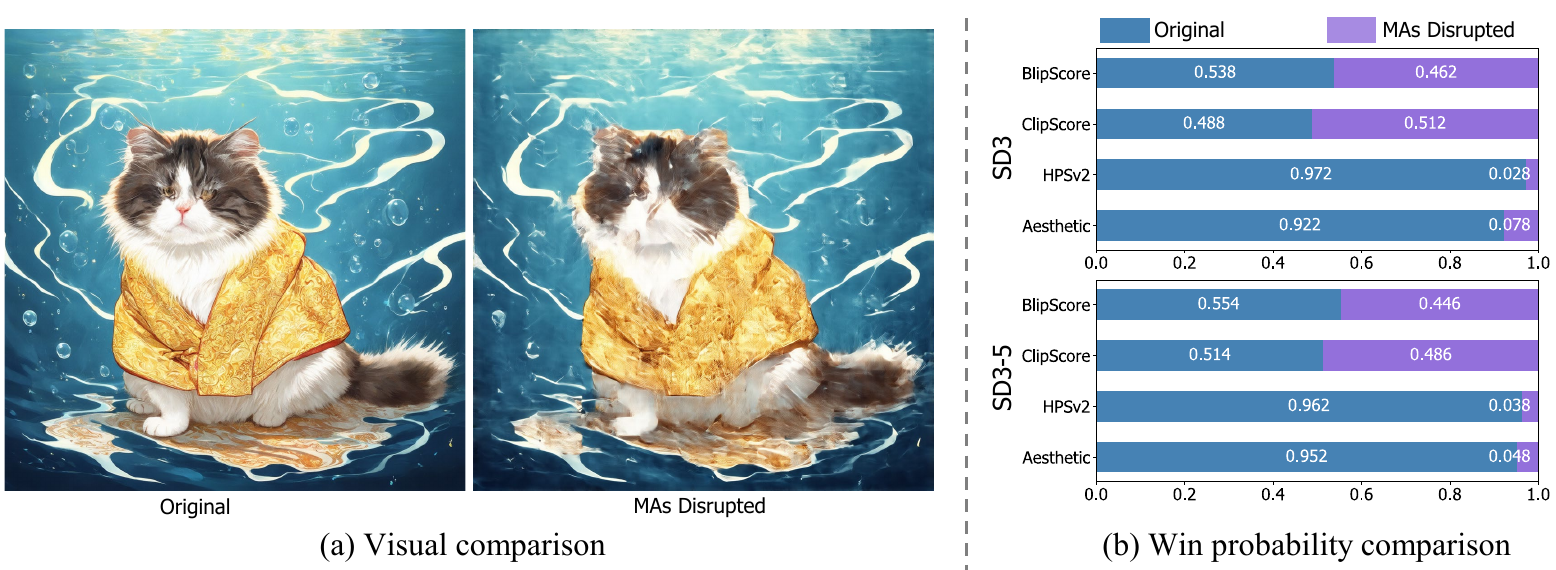}
  \caption{\textbf{Comparison of the original and Massive Activations (MAs) disrupted models.} (a) Sampling results comparison between the original and MAs-disrupted models for SD3. (b) Win probability comparison for different models where we evaluate the model from two perspectives: \textbf{Prompt Alignment} (Blipscore and Clipscore) and \textbf{Local Detail Quality} (HPSv2.1 and Laion-Aesthetics). Disrupting massive activations markedly degrades the fidelity of fine-grained details in the generated images while exerting minimal impact on semantic content.}
  \vspace{-5mm}
  \label{figure:f5}
\end{figure}

Previous works~\citep{darcet2023vision} have characterized massive activations in ViTs, showing that they primarily arise in specific tokens (e.g., background tokens) and serve to encode global information. In contrast, massive activations in DiTs occur across all spatial tokens. This fundamental difference naturally raises a key research question: \textit{What role do massive activations play in DiTs?} To address this question, we perform activation intervention ~\citep{sun2024massive} to examine how massive activations influence the behavior of DiTs. Specifically, we manually disrupt the massive activation values in a single layer and then propagate the modified hidden state through the remaining DiT blocks.
The results are presented in~\Cref{figure:f5}. We provide the full activation intervention setting, including original, Non-MAs disrupted, and MAs-disrupted in~\Cref{sup:intervention}.

\paragraph{Massive activations have minimal impact on semantic content.}
\vspace{-5pt}
We observe that the images generated by the MAs-disrupted model still preserve global semantics such as object identity, color composition, and overall layout, remaining consistent with those from the original model (\Cref{figure:f5}(a)).
Moreover, the MAs-disrupted model maintains comparable prompt alignment metrics, achieving similar Blipscore~\citep{li2022blip} (0.462 vs.\ 0.538) and Clipscore~\citep{radford2021learning} (0.512 vs.\ 0.488) win probabilities relative to the original output (\Cref{figure:f5}(b)).
These results indicate that the inherent massive activations exert minimal influence on the overall semantic content in the visual generation process of DiTs. This finding is consistent with the characteristic described in~\Cref{main:characteristics}, where the input text embedding has negligible effect on massive activations.

\paragraph{Massive activations play a key role in local detail synthesis.}
\vspace{-5pt}
More importantly, it can be easily observed that the fine-grained local details, including textures and subtle object parts (e.g., eyes and hair), are markedly degraded when massive activations are disrupted. Moreover, the MAs-disrupted model attains much lower win probabilities (0.028 on HPSv2.1 and 0.078 on Laion-Aesthetics) than the original model on the local detail quality metric, underscoring the crucial role of massive activations in fine-grained detail synthesis. 

\noindent
In combination with the characteristics of MAs described in~\Cref{main:characteristics}, we summarize two key findings:  
(1) MAs are mainly shaped by the input timestep embedding, and  
(2) they are crucial for local detail synthesis.  
These findings are consistent with the generative dynamics of diffusion models~\citep{ho2020denoising,rombach2022high}:  
the timestep embedding $t$ encodes the noise level and generation stage, with large $t$ guiding coarse structure reconstruction and small $t$ driving fine-grained refinement.  
As sampling proceeds from $T$ to $0$, $t$ modulates the residual scaling factor $\alpha_k$, progressively amplifying massive activations (\Cref{figure:f4}(b)), which in turn orchestrate the detail synthesis process in DiTs.

\subsection{Detail Guidance for Diffusion Transformers}

% Based on these observations, we hypothesize that DiTs learn to assign massive activations across all spatial tokens, where these activations are exploited to encode and refine the local details of each token. 
% During sampling, the timestep embedding regulates their magnitude, thereby enabling adaptive control over fine-grained detail synthesis. 
% As the sampling process progresses, the model progressively amplifies the massive activations (\Cref{figure:f4}(b)), which in turn drive the generation of fine-grained local textures, ultimately yielding high-quality visual outputs.

Based on these findings, we make the following hypothesis: during training, DiT learns to assign massive activations to \textit{all} spatial tokens to \textit{drive} fine-grained local detail synthesis of each token, and uses timestep embeddings to \textit{modulate} massive activations, thereby adaptively \textit{orchestrating} the detail synthesis process throughout generation. 

Motivated by these insights, we seek a \textit{concise and effective} approach to exploit the capacity of MAs for enhancing fine-grained detail synthesis in DiTs. Accordingly, we propose a MAs-driven, training-free self-guidance strategy, termed Detail Guidance (DG). Our approach draws inspiration from the self-guidance mechanism~\citep{karras2024guiding}, which guides the base model with a deliberately degraded ``bad'' version. Different from them, we construct the ``bad'' model by explicitly degrading its capacity for local detail synthesis.

% \paragraph{Detail Guidance (DG).}
% \vspace{-5pt}
Formally, let $D_\theta$ be the original pretrained DiT model and $z_t^k \in \mathbb{R}^{C \times H \times W}$ be the hidden state output of \(k\)-th block. We disrupt the massive activations in $z_t^k$ by masking (zeroing out) the corresponding dimensions to the massive activations and then propagate the modified hidden state $\tilde{z}_t^k$ through the remaining blocks. By disrupting the massive activations (drivers of local detail), we build a degraded model $D_{\theta,m}$ that produces detail-deficient outputs, where $m$ is the disrupted layer depth. Leveraging this degraded ``detail-deficient'' model \(D_{\theta,m}\), we formulate our Detail Guidance (DG) following the diffusion self-guidance mechanism~\citep{karras2024guiding}:
\begin{equation}
\hat{D}_{\theta}(z_t, t, c) =
D_{\theta}(z_t, t, c) + w \big( D_{\theta}(z_t, t, c) - D_{\theta, m}(z_t, t, c) \big)
\end{equation}
where $w$ controls the strength of the detail guidance. Our approach requires no extra training and can be directly applied to mostly pretrained DiT models (~\Cref{tab:SD,tab:flux}).

\paragraph{Integration with CFG.}
\vspace{-5pt}
Classifier-free guidance (CFG)~\citep{ho2022classifier} is a standard technique that enhances semantic alignment by extrapolating between conditional and unconditional predictions. Our DG method is naturally complementary to CFG: whereas CFG strengthens semantic fidelity, DG explicitly enhances local detail quality. 
The combined guidance is expressed as
\begin{equation}
\begin{split}
\hat{D}_{\theta}(z_t, t, c) &=
D_{\theta}(z_t, t, c) +
\lambda \big( D_{\theta}(z_t, t, c) - D_{\theta}(z_t, t) \big) \\
&\quad + w \big( D_{\theta}(z_t, t, c) - D_{\theta, m}(z_t, t, c) \big)
\end{split}
\end{equation}
where $\lambda$ and $w$ are the guidance scales of CFG and DG, respectively. 
% Fundamentally, DG operates in a self-supervised manner: it enhances local detail synthesis by leveraging intrinsic model and data cues without requiring additional labels or training. 
% In contrast, CFG follows a supervised paradigm that relies on externally provided prompt information to promote semantic fidelity and overall visual quality.

\begin{figure}[tp]
  \centering
  \includegraphics[width =\linewidth]{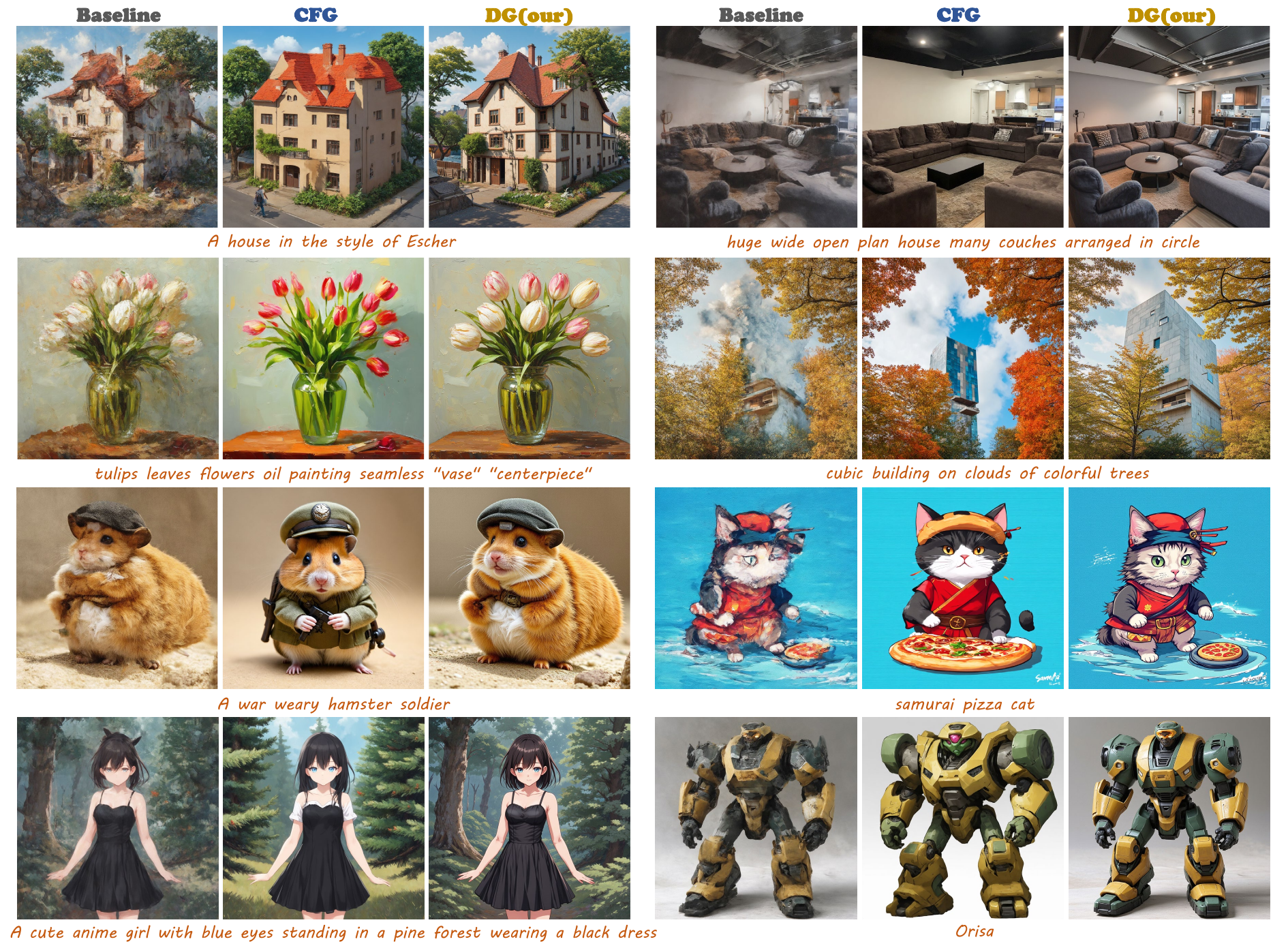}
  \caption{\textbf{Visual results of Detail Guidance (DG) on SD3.}  
    Baseline denotes visual outputs without CFG.
DG produces high-quality images with improved fine-grained details compared to Baseline. The CFG output is included as a reference for better comparison of detail quality.
    }
  \vspace{-5mm}
  \label{figure:f6}
\end{figure}

% \begin{figure}[tp]
%   \centering
%   \includegraphics[width =\linewidth]{img/f6_v1.pdf}
%   \caption{\textbf{Detail Guidance for SD3.} 
%   The top row shows the classifier-free guidance (CFG) output of SD3, the bottom row presents the CFG output with our Detail Guidance (DG) applied. Our DG further enhances the local details of the CFG output.}
%   \vspace{-5mm}
%   \label{figure:f6}
% \end{figure}

\section{Experiments}
\subsection{Experimental Setup}
\noindent\textbf{Model Variants.}
As our approach merely modifies internal hidden states, it can be readily applied to most pretrained DiTs without additional training or tuning.  
We evaluate DG on three representative text-to-image DiTs, SD3-Medium~\citep{esser2024scaling} (SD3), SD3.5-Large~\citep{esser2024scaling} (SD3.5), and Flux~\citep{Flux}.
To comprehensively assess its effectiveness, we test DG under two settings: Conditional (Cond) generation without CFG and CFG generation. The default generated image size is 1024x1024. 
% We provide the results of Flux model in~\Cref{sup:dg_flux}. 
Further implementation details are provided in~\Cref{sup:implementation}.

% We evaluate our strategy mainly on text-to-image tasks.
% For text-to-image generation, we adopt the recent SD3-medium and SD3.5-large models~\citep{esser2024scaling} as baselines, and assess our method under two configurations: Conditional (Cond) and Classifier-Free Guidance (CFG).
% For the class-conditional setting, we conduct experiments on ImageNet generation using the DiT-XL/2 model~\citep{peebles2023scalable}.
% % , and assess our guidance method under three configurations: unconditional, conditional, and classifier-free guidance (CFG).
% To ensure consistency with the original frameworks, all experiments employ each model’s default diffusion sampler and default CFG settings. Additional details on implementation are provided in the Appendix.

\noindent\textbf{Datasets and Evaluation Metric.}
We assess our method on two standard benchmarks: the Pick-a-Pic “test unique” split~\citep{kirstain2023pick} and HPSv2.1~\citep{wu2023human}.
To quantify prompt alignment, we compute Clipscore~\citep{radford2021learning} and Blipscore~\citep{li2022blip}.
To evaluate the fidelity of fine-grained local details, we adopt HPSv2.1~\citep{wu2023human} and Laion-Aesthetics~\citep{schuhmann2022laionaesthetics} as quality metrics. Further evaluation details are provided in~\Cref{sup:evaluation}.

\begin{figure}[tp]
  \centering
  \includegraphics[width =\linewidth]{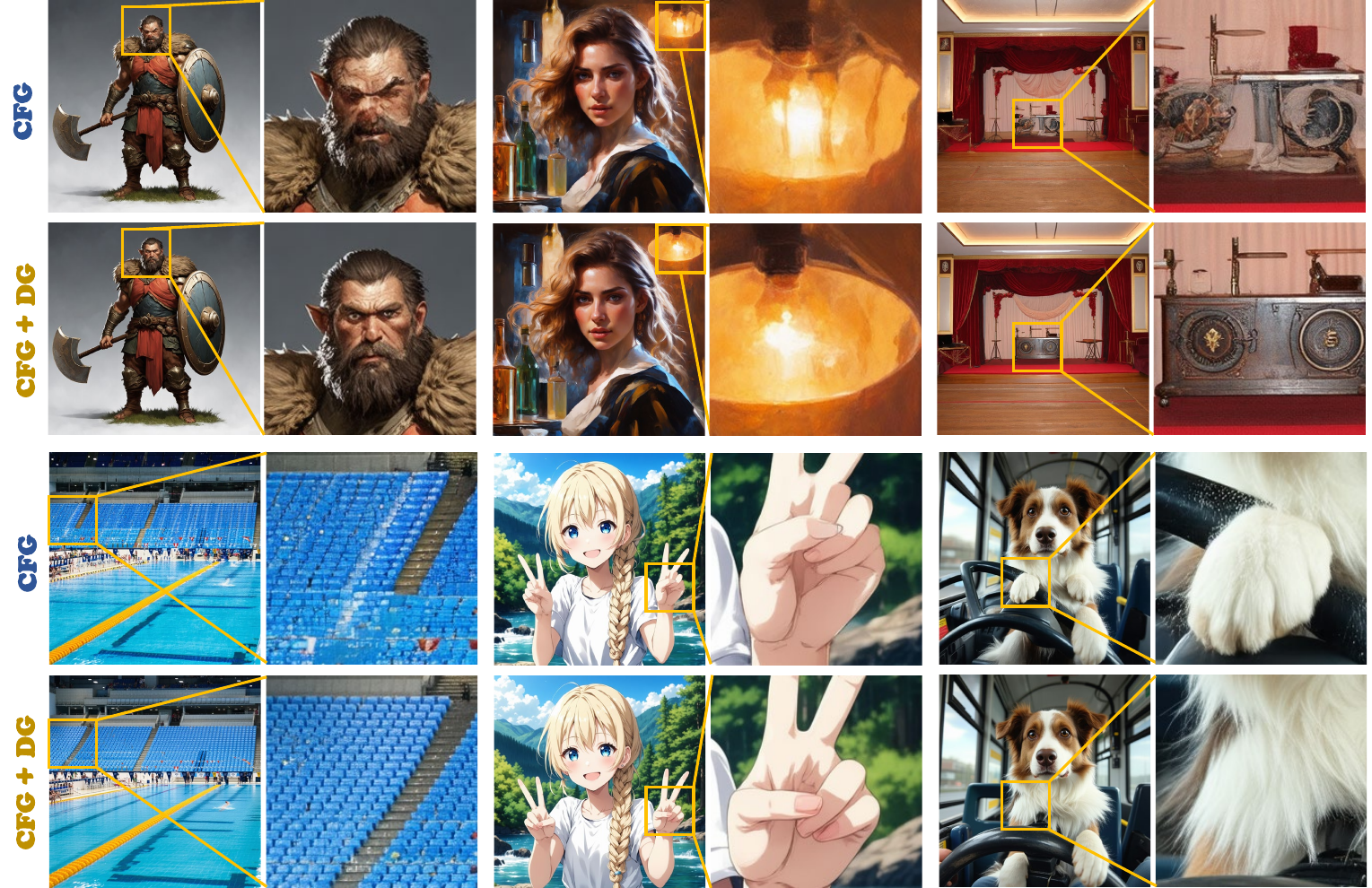}
  \vspace{-5mm}
  \caption{\textbf{Integration with CFG.} Rows 1-2: SD3; Rows 3-4: SD3.5. Incorporating DG into Classifier-Free Guidance (CFG) yields outputs with markedly richer and more refined local details.}
  \vspace{-3mm}
  \label{figure:f7}
\end{figure}

\begin{table*}[t]
\setlength\tabcolsep{14pt}
\centering
\resizebox{0.9\linewidth}{!}{
\begin{tabular}{c|c|c|c|c|c|c}
\hline \multirow{2}{*}{Model} & \multirow{2}{*}{Type} & \multirow{2}{*}{DG} & \multicolumn{2}{c|}{Prompt Alignment} & \multicolumn{2}{c}{Detail Quality} \\
\cline { 4 - 7 } & &  & Clipscore & Blipscore & HPSv2.1 & Aesthetic \\
\hline \multirow{4}{*}{SD3}& \multirow{2}{*}{Cond}& $\times$ & 22.11 & 66.74 & 21.84 &  5.58 \\
 && $\checkmark$ & \textbf{24.15} & \textbf{76.52} & \textbf{28.65} & \textbf{6.01}  \\
\cline { 2 - 7 }  &\multirow{2}{*}{CFG} & $\times$ & \textbf{26.64} & \textbf{87.01} & 28.23 & 5.80 \\
 % && $\checkmark$ & 26.25 & 86.32 & \textbf{29.87} &\textbf{5.94} \\
 && $\checkmark$ & 26.25 & 86.32 & \textbf{29.87} &\textbf{6.03} \\
\hline \multirow{4}{*}{SD3.5}& \multirow{2}{*}{Cond}& $\times$ & 24.90 & 70.09 & 23.65 &  5.94\\
 && $\checkmark$ & \textbf{26.01} & \textbf{83.66} & \textbf{29.23} & \textbf{6.16} \\
\cline { 2 - 7 }  &\multirow{2}{*}{CFG} & $\times$ & 27.67 & \textbf{92.62} & 29.9 & 6.01\\
 && $\checkmark$ & \textbf{27.68} & 91.61 & \textbf{30.7} & \textbf{6.18} \\
% \hline \multirow{4}{*}{Flux}& \multirow{2}{*}{Cond}& $\times$ & 22.09 & 57.60 & 19.33 &  5.50\\
%  && $\checkmark$ & \textbf{25.69} & \textbf{80.55} & \textbf{27.88} & \textbf{6.13} \\
% \cline { 2 - 7 }  &\multirow{2}{*}{CFG} & $\times$ & 27.04 & \textbf{87.76} & 29.16 & 5.96\\
%  && $\checkmark$ & \textbf{} &  & \textbf{} & \textbf{} \\
\hline
\end{tabular}}
\vspace{-2mm}
\caption{\textbf{Quantitative comparison on dataset Pick-
a-Pic.}
% We evaluate our method on two generation settings: Conditional (Cond) and Classifier-Free guidance (CFG).
Our DG strategy brings substantial improvements on detail quality for both settings, demonstrating its effectiveness in enhancing visual details. See~\Cref{sup:dg_flux} for the evaluation of DG on Flux.
The best highlights in bold.}
\label{tab:SD}
\vspace{-3mm}
\end{table*}

\vspace{-2mm}
\subsection{Main results on visual generation}
\vspace{-1mm}
% To evaluate our strategy, we conduct text-to-image generation with model SD3, SD3-5 and Flux on dataset Pick-a-Pic~\citep{kirstain2023pick} and HPSv2.1~\citep{wu2023human}. 
% % and class-conditional generation on ImageNet 256$\times$256~\citep{deng2009imagenet}.

\noindent\textbf{Evaluation of DG.}
\Cref{tab:SD,tab:flux} reports the quantitative results of Detail Guidance (DG) on three pretrained DiTs. 
DG achieves substantial improvements in both prompt alignment and detail quality, (e.g., Blipscore from 70.09 to 83.66 and Aesthetic from 5.94 to 6.16 on SD3.5). 
Qualitative results in~\Cref{figure:f6} further confirm that DG effectively enhances fine-grained local details while faithfully preserving the overall image structure. We provide qualitative results on SD3.5 and Flux in~\Cref{sup:qualitative}.  
Moreover, we report experiments on ImageNet 256×256 in~\Cref{sup:class}, showing that our DG strategy also enhances visual quality in class-conditional generation tasks.

\noindent\textbf{DG versus CFG.}
From~\Cref{tab:SD}, DG yields higher detail-quality scores (e.g., Aesthetic 6.01 vs.\ 5.80 for SD3), whereas CFG achieves stronger prompt alignment. 
As illustrated in~\Cref{figure:f6}, DG produces outputs with richer local textures, while CFG excels at semantic alignment. 
These results indicate that DG primarily enhances local detail synthesis, while CFG strengthens semantic alignment.

\noindent\textbf{Integrating DG with CFG.}
DG integrates seamlessly with CFG, consistently improving detail quality as shown in~\Cref{tab:SD}. 
Visual comparisons in~\Cref{figure:f7} highlight that the combined strategy further refines fine-grained details, yielding higher overall image quality.
We provide more visual results of SD3 and SD3.5 in~\Cref{sup:combination}.

\noindent\textbf{Comparison with other guidance.}
We also compare DG with other guidance strategies(e.g., CFG++ and APG) on HPSv2.1 (\Cref{tab:guidance}). 
DG requires no training while still achieving the highest Aesthetic score and competitive HPSv2.1 performance.
When combined with CFG, our method establishes a new state of the art, underscoring its effectiveness in enhancing visual quality.

\begin{table*}[t]
\setlength\tabcolsep{9pt}
\centering
\resizebox{0.9\linewidth}{!}{
\begin{tabular}{l|l|c|c|c|c|c|c}
\hline \multirow{2}{*}{Train}&\multirow{2}{*}{Method} &\multicolumn{5}{c|}{HPSv2.1} &\multirow{2}{*}{Aesthetic} \\
\cline { 3 - 7 } && Anime& Concept& Painting & Photo & Avg. &\\
\hline \multirow{4}{*}{$\checkmark$}&CFG &31.34&30.62&30.98&28.01&30.24 & 5.93\\
&APG &30.76&29.98&30.24&26.86&29.46 & 5.89\\
&CFG++ &31.58&30.32&30.95&27.24&30.02& 5.91\\
&CFG-Zero &\underline{31.64}& \underline{31.05}&\textbf{31.35}&28.25&\underline{30.57}& 6.07\\
\hline $\times$ &DG (Ours)&31.14&30.17&30.05&\underline{28.70}&30.14&\textbf{6.14} \\
\hline $\checkmark$ &CFG+DG (Ours)&\textbf{32.23}&\textbf{31.11}&\underline{31.27}&\textbf{29.21}&\textbf{30.96}& \underline{6.13}\\
\hline
\end{tabular}}
\captionof{table}{\textbf{Evaluation of different guidance on dataset HPSv2.1 with SD3.} Train means whether need to train an unconditional branch. The best highlights in bold, while the second best is underlined.}
\label{tab:guidance}
\vspace{-7pt}
\end{table*}

\begin{figure}[tp]
  \centering
  \includegraphics[width =\linewidth]{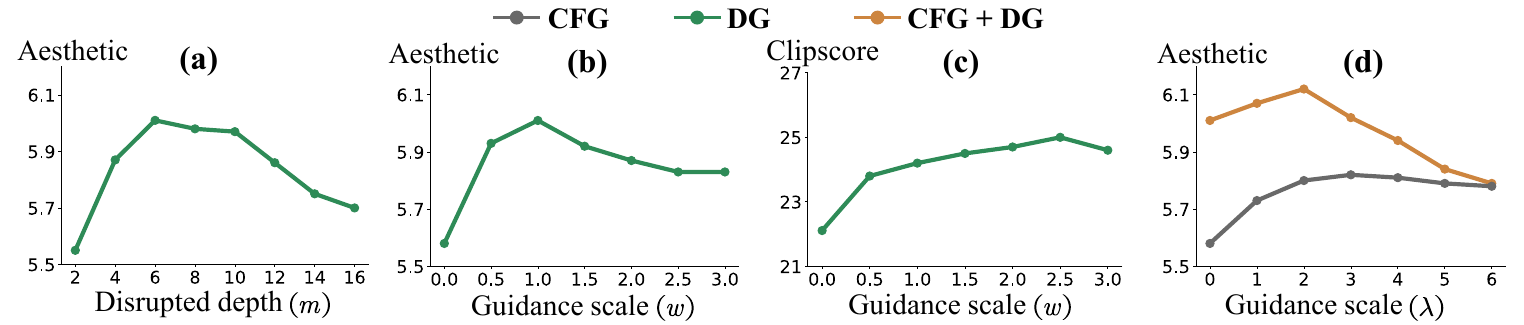}
  \caption{\textbf{Investigations of disrupted depth $m$, scales $\lambda$ and $w$ for SD3.}}
  \vspace{-6mm}
  \label{figure:f8}
\end{figure}

\vspace{-2mm}
\subsection{Further analysis}
\noindent\textbf{Disrupted depth $m$.}
We examine the effect of the disrupted depth $m$ of massive activations, as shown in~\Cref{figure:f8}(a).
Our DG strategy achieves the best performance when applied to intermediate blocks 
(e.g., $m$ ranging from 4 to 10).
We hypothesize that early blocks mainly contain heavy noise and lack even coarse image structures, making disruption there uninformative,
while applying disruption in late blocks occurs too close to the final output and thus has minor impact on generation.
Based on these observations, we primarily perturb massive activations in the intermediate layers and set the default $m=6$ for the SD3 model.
The configurations for SD3.5 and Flux are provided in~\Cref{sup:implementation}.

\noindent\textbf{Guidance scales $w$ and $\lambda$.}
We present the quantitative results across different scales in~\Cref{figure:f8}.
Our DG consistently achieves stable and high Aesthetic (AES) scores (\Cref{figure:f8}(b))
and CLIPScore (\Cref{figure:f8}(c)).
When combined with CFG, it further boosts AES performance (\Cref{figure:f8}(d)).
These results highlight the effectiveness and robustness of our approach in enhancing fine-grained details.
% Moreover, DG integrates seamlessly with CFG, enabling joint improvements in local-detail fidelity and prompt alignment.

\noindent\textbf{User study.}
We conduct a user study to evaluate the benefits of our DG strategy from three key aspects: prompt alignment, color consistency, and detail preservation(see~\Cref{sup:user}).
The results show that our DG strategy yields substantial improvements in
color consistency and detail preservation,
demonstrating its effectiveness in enhancing fine-grained visual details.

\vspace{-1mm}
\section{Conclusion}
\vspace{-2mm}
In this paper, we systematically investigate an intriguing phenomenon in DiTs, termed \emph{Massive Activations} (MAs).
We find that these activations emerge across all spatial tokens and that their distribution is shaped by the input timestep embeddings.
Our further analysis demonstrates that these activations are critical for local detail synthesis in DiTs. We interpret them as drivers of local detail information whose magnitude is dynamically modulated by timestep embeddings, thereby orchestrating detail synthesis during the DiT generation process.
Building on these insights, we propose \emph{Detail Guidance} (DG), a MAs-driven, training-free self-guidance strategy to explicitly enhance local detail synthesis. Our DG can be seamlessly combined with CFG, enabling joint enhancement of detail fidelity and prompt alignment. Extensive experiments demonstrate the effectiveness of our approach in improving fine-grained detail synthesis.

\bibliography{iclr2026_conference}
\bibliographystyle{iclr2026_conference}

\newpage
\appendix

\section*{\textcolor{citecolor}{Appendix}\\
Massive Activations are the Key to Local Detail Synthesis in Diffusion Transformers
}
% \section{Massive activations in DiTs}

\section{Diffusion Transformer Architecture}
\label{sup:dit}
\begin{figure}[ht]
  \centering
  \includegraphics[width =0.35\linewidth]{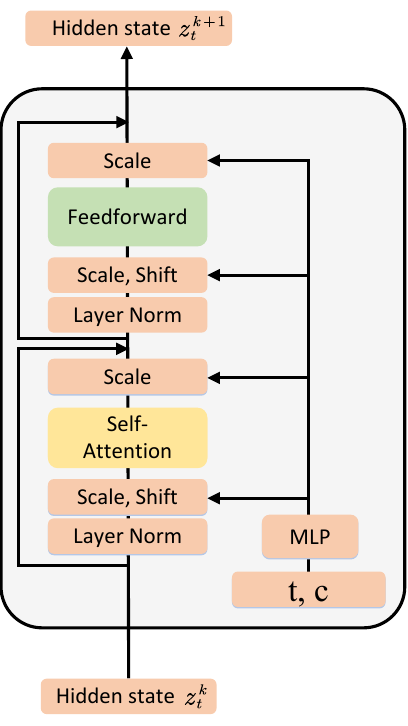}
  \caption{\textbf{Illustration of the architecture of DiT block $D_k$.}
  % DiTs Only achieve performances of approximately 40 on PCK@0.10 metric in the Spair-71k dataset, while the performance of SD2-1(Unet) is about 60.
  }
  \vspace{-4mm}
  \label{figure:dit}
\end{figure}
We present the architecture of a DiT block in~\Cref{figure:dit}. 
Each block consists of three key components: an \(\operatorname{AdaLN}\) layer, a \(\operatorname{Self\text{-}Attention}\) layer, and a \(\operatorname{Feedforward}\) layer. 
The \(\operatorname{AdaLN}\) layer encodes the input timestep \(t\) and the additional conditioning information \(c\) (e.g., class or text embedding) into channel-wise scale and shift parameters \(\gamma_k\) and \(\beta_k\). 
It then performs Adaptive Layer Normalization (AdaLN) on the hidden state \(z_t^k\):
\begin{equation}
\hat{z}_t^k = \bigl(1+\gamma_k\bigr)\,\operatorname{LayerNorm}(z_t^k) + \beta_k,
\label{AdaLN}
\end{equation}
where $\gamma_k, \beta_k$ are regressed by the MLP networks of AdaLN layer conditioned on input timestep embedding $t$ and text embedding $c$:
\begin{gather}
\gamma_k, \beta_k, \alpha_k = \operatorname{MLP}_k(t, c)
\label{regress}
\end{gather}
where $\alpha_k$ scales the $k$-th residual connection.

Next, \(\hat{z}_t^k\) is processed by the \(\operatorname{Self\text{-}Attention}\) layer to produce an intermediate feature representation.  
A second adaptive layer normalization is then applied before passing this intermediate feature to the \(\operatorname{Feedforward}\) layer, which outputs the updated hidden state.  
Finally, a residual connection combines the input and the transformed features to produce the block output:
\begin{equation}
z_t^{k+1} = z_t^{k} + \alpha_k \,D_k(z_t^k, t),
\label{sup:block1}
\end{equation}
% where \(\alpha_k\) is the residual scaling factor predicted by the built-in AdaLN layer.
\newpage
\section{Layer properties of Massive activations}
\label{sup:layer}
\begin{figure}[htp]
  \centering
  \includegraphics[width =0.33\linewidth]{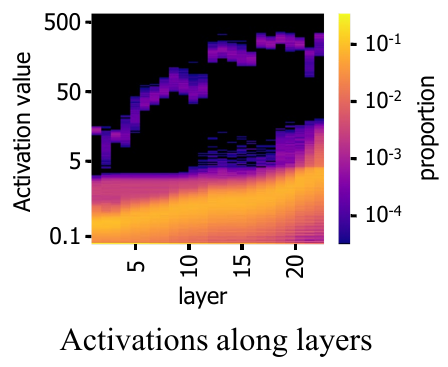}
  \caption{\textbf{Layer properties of MAs in SD3.} Massive activations in SD3 occur in all layers.}
  \vspace{-2mm}
  \label{sup:sd3_layer}
\end{figure}

\begin{figure}[htp]
  \centering
  \includegraphics[width =0.33\linewidth]{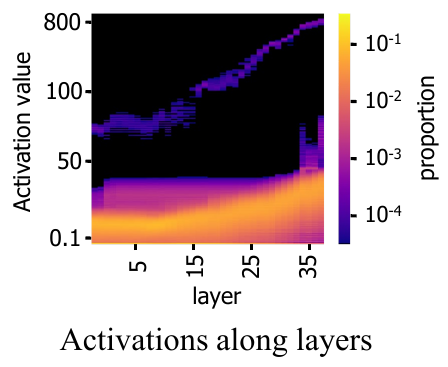}
  \caption{\textbf{Layer properties of MAs in SD3.5.} Massive activations in SD3.5 occur in all layers.}
  \vspace{-2mm}
  \label{sup:sd35_layer}
\end{figure}

\begin{figure}[htp]
  \centering
  \includegraphics[width =0.33\linewidth]{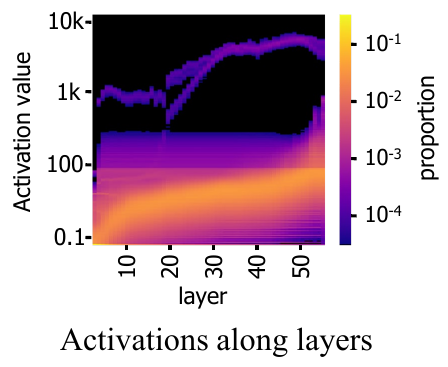}
  \caption{\textbf{Layer properties of MAs in Flux.} Massive activations in Flux occur in all layers.}
  \vspace{-2mm}
  \label{sup:flux_layer}
\end{figure}

In this section, we examine the layer-wise characteristics of MAs in SD3, SD3.5, and Flux.  
The results are shown in~\Cref{sup:sd3_layer,sup:sd35_layer,sup:flux_layer}, revealing that massive activations consistently occur throughout all layers in these DiT models.

\section{Timestep vs Text impact on Massive activations}
\label{sup:timestep}

\begin{figure}[t] % 使用 [p!] 浮动参数，强烈建议 LaTeX 将此图放在一个单独的页面上
  \centering

  % --- 第一张图 ---
  \begin{minipage}{\linewidth}
    \centering
    \includegraphics[width=\linewidth]{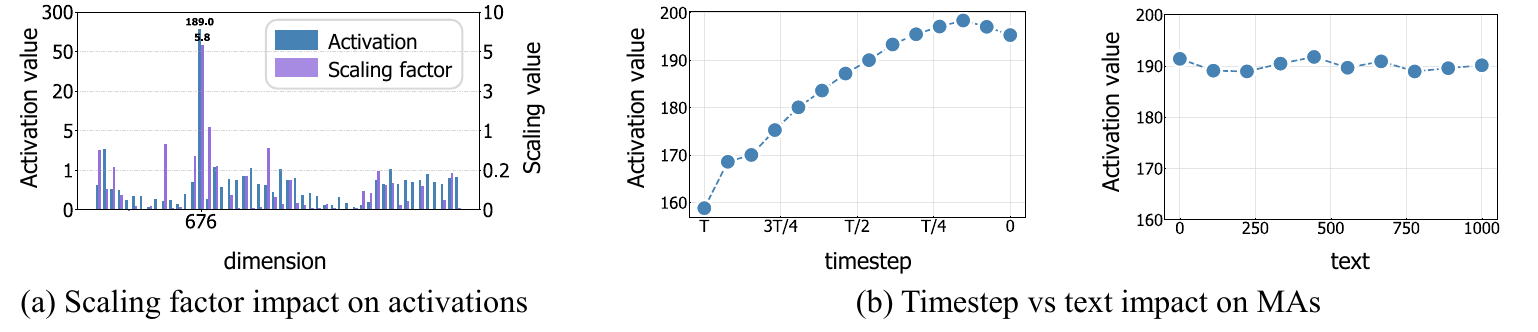}
    \caption{\textbf{Impact of the input timestep and text on Massive Activations (MAs) in SD3.5.} The input timestep $t$ plays a dominant role for massive activation: the magnitude of massive activations increases steadily as $t$ decreases from $T$ to $0$.}
    \vspace{-2mm}
    \label{figure:a1}
  \end{minipage}

  \vspace{8mm} % 在第一张图和第二张图之间添加一些固定的垂直间距

  % --- 第二张图 ---
  \begin{minipage}{\linewidth}
    \centering
    \includegraphics[width=\linewidth]{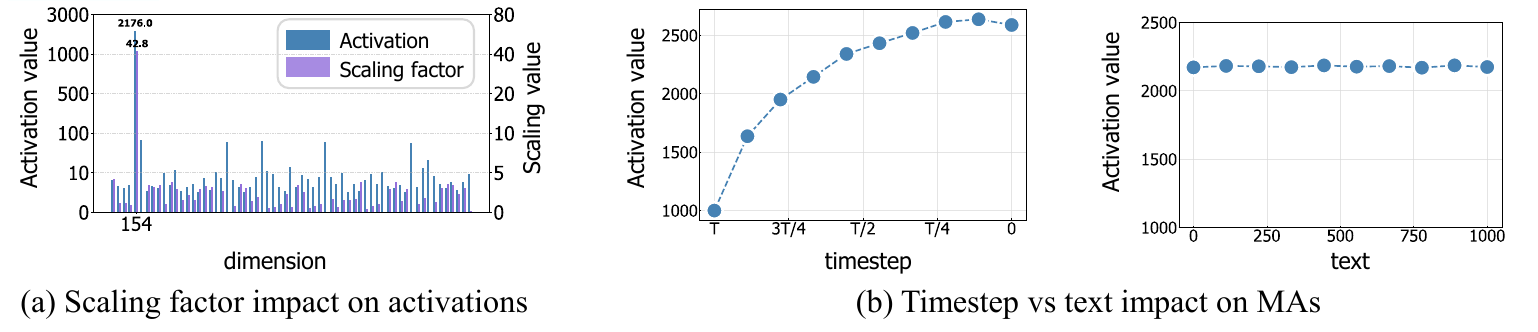}
    \caption{\textbf{Impact of the input timestep and text on Massive Activations (MAs) in Flux.} The input timestep $t$ plays a dominant role for massive activation: the magnitude of massive activations increases steadily as $t$ decreases from $T$ to $0$.}
    \vspace{-2mm}
    \label{figure:a2}
  \end{minipage}

  \vspace{8mm} % 在第二张图和第三张图之间添加一些固定的垂直间距

  % --- 第三张图 ---
  \begin{minipage}{\linewidth}
    \centering
    \includegraphics[width=\linewidth]{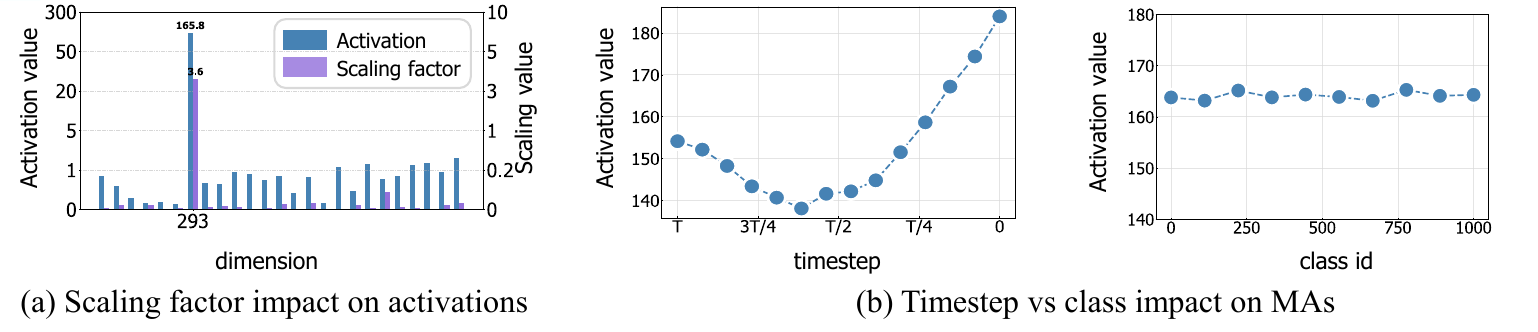}
    \caption{\textbf{Impact of the input timestep and text on Massive Activations (MAs) in DiT-XL.} The input timestep $t$ plays a dominant role for massive activation: the magnitude of massive activations increases steadily as $t$ decreases from $T/2$ to $0$.}
    \vspace{-2mm}
    \label{figure:a3}
  \end{minipage}

\end{figure}

We provide additional analysis for the massive activation of SD3-5, Flux and DiT-XL in~\Cref{figure:a1,figure:a2,figure:a3}. As the hidden states $z_{t}^{k+1}$ in DiTs are computed via a residual connection (\Cref{block}):
\begin{equation}
z_{t}^{k+1} = z_{t}^{k}+\alpha_k\mathcal{D}_k(z_{t}^{k}, t)
\label{sup:block}
\end{equation}
where the $\alpha_k$ is the residual scaling factor. We first examine the impact of the residual scaling factor on these activations. Specifically, we visualize the average magnitude of each dimension for activation values and scaling factor values. We observe that the residual scaling factors $\alpha_k$ govern the dimension and values of massive activations in DiTs.

% \begin{figure}[tp]
%   \centering
%   \includegraphics[width =\linewidth]{img/f4_sd35.pdf}
%   \caption{\textbf{Impact of the input timestep and text on Massive Activations (MAs) in SD3-5.} The input timestep $t$ plays a dominant role for massive activation: the magnitude of massive activations increases steadily as $t$ decreases from $T$ to $0$.}
%   \vspace{-2mm}
%   \label{figure:a1}
% \end{figure}

% \begin{figure}[tp]
%   \centering
%   \includegraphics[width =\linewidth]{img/f4_flux.pdf}
%   \caption{\textbf{Impact of the input timestep and text on Massive Activations (MAs) in Flux.} The input timestep $t$ plays a dominant role for massive activation: the magnitude of massive activations increases steadily as $t$ decreases from $T$ to $0$.}
%   \vspace{-2mm}
%   \label{figure:a2}
% \end{figure}

% \begin{figure}[h]
%   \centering
%   \includegraphics[width =\linewidth]{img/f4_v5.pdf}
%   \caption{\textbf{Impact of the input timestep and text on Massive Activations (MAs) in DiT-XL.} The input timestep $t$ plays a dominant role for massive activation: the magnitude of massive activations increases steadily as $t$ decreases from $T/2$ to $0$.}
%   \vspace{-2mm}
%   \label{figure:a3}
% \end{figure}

Furthermore, the residual scaling factor $\alpha_k$ is regressed by the AdaLN layer conditioned on input timestep embedding $t$ and text embedding $c$:
\begin{gather}
\alpha_k = \operatorname{MLP}_k(t, c)
\label{sup:regress}
\end{gather}
Therefore, we next investigate the respective impact of the input timestep $t$ and text $c$ to the massive activations. As shown in~\Cref{figure:a1}(b),~\Cref{figure:a2}(b) and~\Cref{figure:a3}(b), it can be found that the magnitude of MAs is mainly influenced by the input timestep embedding while the input text embedding exerts minimal impact on it.

\section{Activation Intervention for DiTs}
\label{sup:intervention}

To better understand the functional role of Massive Activations (MAs) in the visual generation of Diffusion Transformers (DiTs), we conduct an activation intervention study with three experimental settings:

\begin{itemize}
    \item \textbf{Original.}  
    The pretrained DiT models are used to generate visual outputs without any modification.

    \item \textbf{MAs Disrupted.}  
    We disrupt the massive activations by masking (zeroing out) their corresponding dimensions (e.g., dimension 293 for SD3 in~\Cref{figure:f2}), as MAs consistently occur at fixed dimensions across all spatial tokens.  
    Specifically, we mask the massive-activation dimensions in the block-output hidden state of a single block and propagate the modified state through the remaining DiT blocks.  
    All other configurations (e.g., sampling steps) are kept same to the Original setting to ensure fair comparison.

    \item \textbf{Non-MAs Disrupted.}  
    To provide a rigorous control, we additionally mask an equal number of randomly selected non-Massive dimensions instead of the massive-activation dimensions.  
    This setting verifies that any observed effect arises specifically from disrupting MAs rather than from the masking operation itself.
\end{itemize}

\begin{figure}[t]
  \centering
  \includegraphics[width =0.8\linewidth]{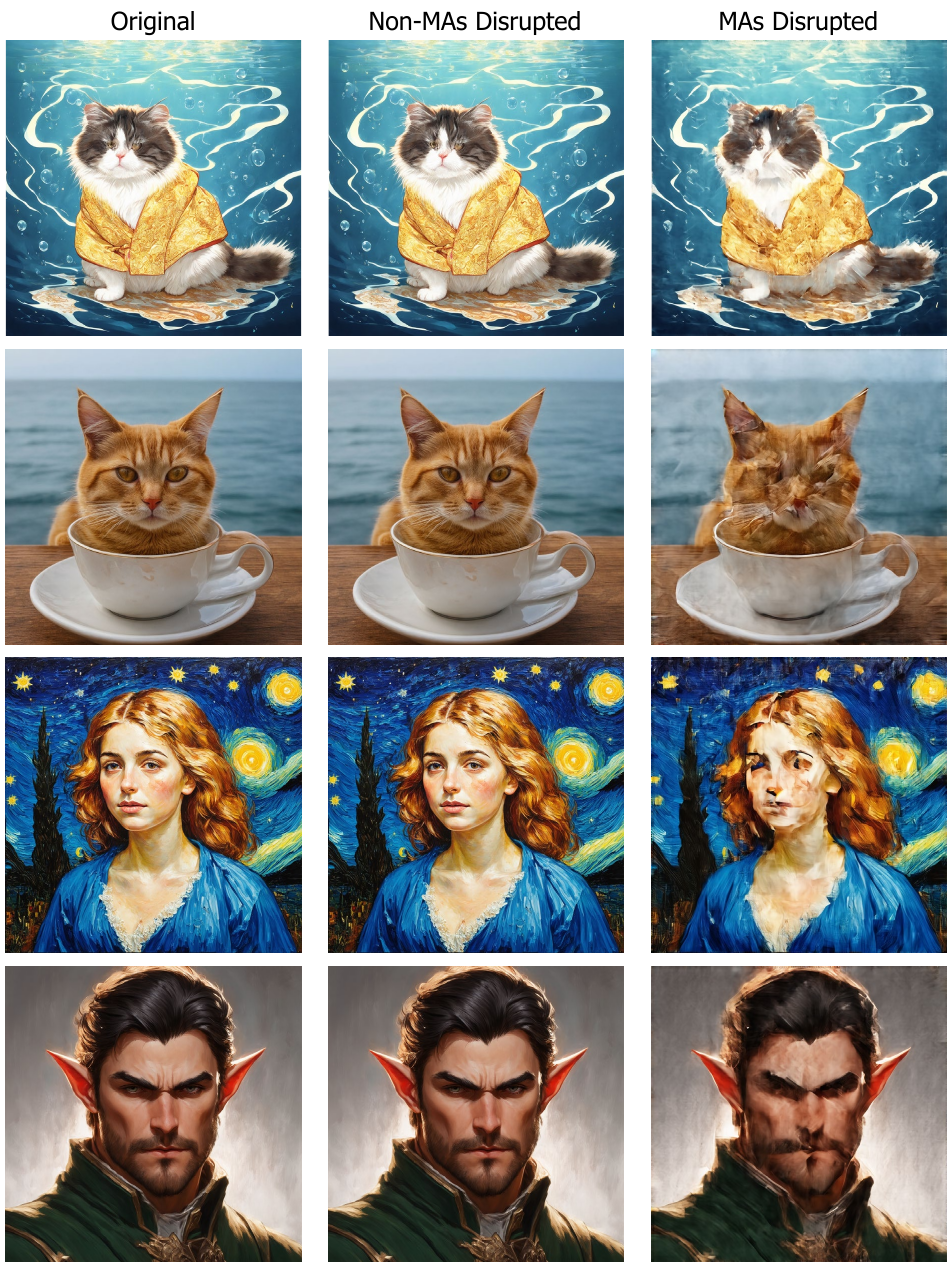}
  \caption{\textbf{Visual comparison of activation intervention.}  
MAs-disrupted models produce images with noticeably degraded local details, whereas non-MAs disruption preserves similar high-quality details with the original outputs.  
These results demonstrate that massive activations are crucial for fine-grained local detail synthesis in DiT generation process.
  }
  \vspace{-4mm}
  \label{figure:MAs}
\end{figure}

We present the results in~\Cref{figure:MAs}.  
It can be observed that disrupting the massive activations in DiTs markedly degrades the fidelity of fine-grained local details, whereas disrupting non-MA dimensions has almost no effect on the generated images.  
These findings indicate that massive activations play a crucial role in driving the synthesis of fine-grained local details during the visual generation process of DiTs.

\section{Evaluation of DG on Flux}
\label{sup:dg_flux}

\begin{table*}[htp]
\setlength\tabcolsep{14pt}
\centering
\resizebox{\linewidth}{!}{
\begin{tabular}{c|c|c|c|c|c|c}
\hline \multirow{2}{*}{Model} & \multirow{2}{*}{Type} & \multirow{2}{*}{DG} & \multicolumn{2}{c|}{Prompt Alignment} & \multicolumn{2}{c}{Detail Quality} \\
\cline { 4 - 7 } & &  & Clipscore & Blipscore & HPSv2.1 & Aesthetic \\
\hline \multirow{4}{*}{Flux}& \multirow{2}{*}{Cond}& $\times$ & 22.09 & 57.60 & 19.33 &  5.50\\
 && $\checkmark$ & \textbf{25.69} & \textbf{80.55} & \textbf{27.88} & \textbf{6.13} \\
\cline { 2 - 7 }  &\multirow{2}{*}{CFG} & $\times$ & 27.04 & \textbf{87.76} & 29.16 & 5.96\\
 && $\checkmark$ & \textbf{27.14} & 86.23 & \textbf{29.25} & \textbf{6.12} \\
\hline
\end{tabular}}
\caption{\textbf{Quantitative performance comparison on Flux.}}
\label{tab:flux}
\vspace{-2mm}
\end{table*}
We further evaluate our DG strategy on the Flux model.
As shown in~\Cref{tab:flux}, our strategy effectively enhances visual quality,
particularly improving fine-grained details.
When combined with CFG, it yields consistent performance gains,
highlighting the robustness and efficacy of our approach.

\section{More Implementation Details}
\label{sup:implementation}
We implement DG on three pretrained large diffusion models: 
\textbf{SD3-Medium}~\citep{esser2024scaling}, 
\textbf{SD3.5-Large}~\citep{esser2024scaling}, 
and \textbf{Flux-dev}~\citep{Flux}. 
Notably, \textbf{Flux} is a CFG-distilled model. 
To evaluate DG independently of CFG, we adopt the \emph{de-distilled} variant from~\citep{Flux}. 
Full experimental settings are provided in~\Cref{tab:config,tab:hyper}.

\noindent\textbf{Configurations for DG.}
For each Diffusion Transformer, we construct a degraded \emph{detail-deficient} model $D_{\theta,m}$ 
by disrupting the dimensions corresponding to massive activations in the $m$-th blocks, 
following the intrinsic activation patterns of each DiT. 
Detailed configurations are summarized in~\Cref{tab:config}.

\noindent\textbf{Hyperparameters setup.}  
All models adopt the default diffusion sampling settings 
(e.g., sampler type and number of steps). 
Specific hyperparameter choices are listed in~\Cref{tab:hyper}.

\noindent\textbf{Computing Resources.}  
All experiments are performed on a single NVIDIA L40S (48\,GB) GPU. 
DG builds the degraded \emph{detail-deficient} model by directly disrupting 
massive activations in hidden states \textbf{without additional training}. 
% and its inference speed is less than that of CFG.

\begin{table*}[htp]
% \vspace{-1mm}
\setlength\tabcolsep{10pt}
\centering
\resizebox{\linewidth}{!}{
\begin{tabular}{l|cc|ccc}
\hline  Model & Blocks N& Hidden size d &Disrupted dimensions&Disrupted depth $m$\\
SD3&24&1536&810&6 \\
SD3-5&38&2432&676&20\\
Flux&57&3072&[154, 1446]&22 \\
\hline
\end{tabular}
}
\caption{\textbf{Configurations of Detail Guidance (DG) for different DiTs.}}
\label{tab:config}
\vspace{-2mm}
\end{table*}

\begin{table*}[htp]
\setlength\tabcolsep{22pt}
\centering
\resizebox{\linewidth}{!}{
\begin{tabular}{c|c|c|c|c}
\hline Model & Guidance Type & sampling step& $\lambda$&$w$ \\
\hline \multirow{3}{*}{SD3}& CFG & 28&4&- \\
& DG &28& -& 1 \\
& CFG+DG &28&3 &1 \\
\hline \multirow{3}{*}{SD3-5}& CFG &28 & 4&- \\
& DG &28& -& 4 \\
& CFG+DG &28& 3& 2\\
\hline \multirow{3}{*}{Flux}& CFG &50& 3.5&- \\
& DG &50& -& 4 \\
& CFG+DG &50& 3 & 2\\
\hline
\end{tabular}}
\caption{\textbf{Hyperparameter setup.}}
\label{tab:hyper}
\vspace{-2mm}
\end{table*}

\section{Evaluation Details}
\label{sup:evaluation}

We evaluate different guidance strategies from two key perspectives: 
\emph{prompt alignment} and \emph{detail quality}. 
Prompt alignment reflects how well the generated image semantically matches the input prompt, 
while detail quality measures the fidelity and richness of fine-grained visual details.

Specifically, we adopt Clipscore~\citep{radford2021learning} and Blipscore~\citep{li2022blip} 
to quantify prompt alignment, and employ HPSv2.1~\citep{wu2023human} and 
Laion-Aesthetics~\citep{schuhmann2022laionaesthetics} as indicators of visual detail quality. 
The details of each metric are as follows.

\noindent\textbf{Clipscore} measures the global semantic consistency between text and image by computing the cosine similarity 
between their CLIP-encoded features. 
We adopt the \textit{clip-vit-large-patch14} version for all experiments.

\noindent\textbf{Blipscore} estimates prompt-image alignment through a fine-grained image-text matching model (BLIP), capturing nuanced semantic relationships beyond global similarity. We use the \textit{blip-itm-large-coco} version to evaluate the model.

\noindent\textbf{HPSv2.1} is a human preference score that provides a perceptual measure of visual realism and aesthetic quality. 
It is widely used to benchmark high-fidelity image synthesis, and we adopt HPSv2.1 for evaluation.

\noindent\textbf{Laion-Aesthetics} predicts aesthetic appeal using a model trained on LAION's large-scale human-rated dataset, 
serving as an automated proxy for human aesthetic assessment. 
% We use the \textit{aesthetic predictor V2.5} for evaluation.

\section{Class-conditional generation}
\label{sup:class}
To evaluate the robustness of our Detail Guidance (DG) strategy,  
we perform class-conditional generation on the ImageNet 256×256 dataset by applying DG to the pretrained DiT-XL/2 model~\citep{peebles2023scalable}.

\begin{table*}[htp]
    \centering
    \setlength\tabcolsep{12pt}
    \resizebox{0.7\linewidth}{!}{
    \begin{tabular}{c|c|c|c|c|c}
    \hline Type & DG & FID $\downarrow$ & IS $\uparrow$ & Prec. $\uparrow$ & Rec. $\uparrow$ \\
    \hline  \multirow{2}{*}{Uncond}& $\times$ & 16.95 & 105.64 & 0.61 & 0.76 \\
     & $\checkmark$ & 9.68 & 122.22 & 0.66 & 0.67 \\
    \hline  \multirow{2}{*}{Cond} & $\times$ & 9.52 & 122.79 & 0.66 & 0.63 \\
     & $\checkmark$ & 5.77 & 179.26 & 0.78& 0.55 \\
    % \hline  \multirow{2}{*}{CFG}& $\times$ & 2.26 & 279.91 & 0.82 & 0.57 \\
    %  & $\checkmark$ & - & - & - & - \\
    \hline
    \end{tabular}}
    \captionof{table}{\textbf{Performance comparison on dataset ImageNet $256 \times 256$.} Prec: Precision, Rec: Recall.}
   \label{tab:imagenet}
   \vspace{-5pt}
\end{table*}

For DiT-XL/2, we set the disrupted depth $m=7$.  
We assess DG under both unconditional and conditional generation settings, with results reported in~\Cref{tab:imagenet}.  
DG delivers consistent performance improvements in both settings, demonstrating the robustness of our guidance strategy.

\section{User study}
\label{sup:user}

\begin{figure}[htp]
  \centering
  \includegraphics[width =\linewidth]{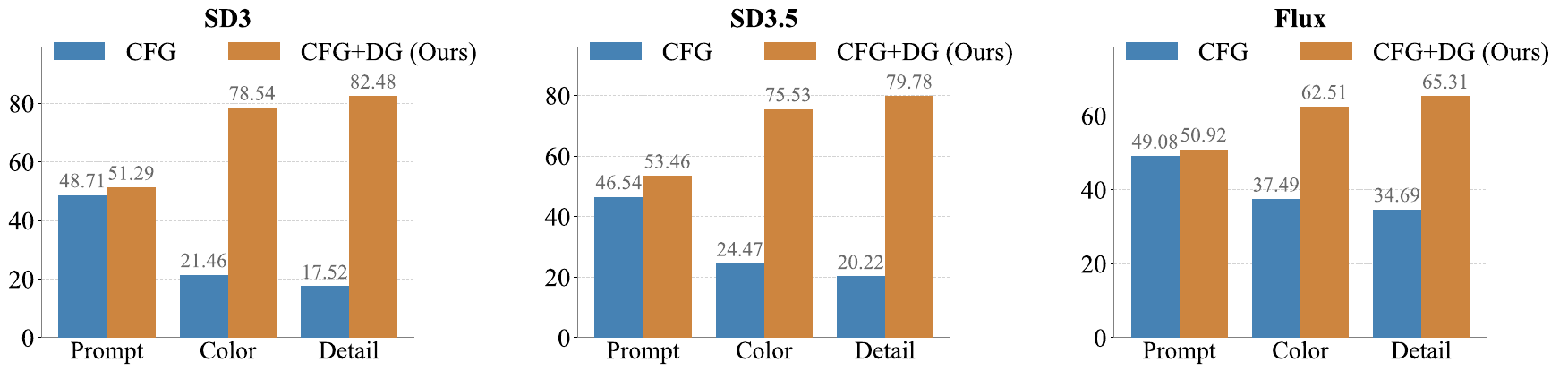}
  \caption{\textbf{User study on SD3, SD3.5, and Flux.}
We report the win rates comparing CFG with our method.
  }
  \vspace{-4mm}
  \label{figure:user}
\end{figure}
We conduct a user study to evaluate the benefits of our DG strategy from three key aspects:
\textbf{prompt alignment}, \textbf{detail preservation}, and \textbf{color consistency}.
Prompt alignment measures how well the generated images match the input prompts.
Detail preservation reflects the fidelity of fine-grained visual details,
while color consistency captures the naturalness and realism of colors.

For each model, 20 annotators compared 100 pairs of images produced by
\textbf{CFG} and \textbf{CFG + DG} with respect to these criteria.
We report the averaged win rates in~\Cref{figure:user},
which show that our approach yields significant improvements across all metrics,
particularly for \emph{Detail} and \emph{Color}.

% \section{Use of Large Language Models (LLMs)}
% \label{sup:LLM}
% In preparing this manuscript, we occasionally used Large Language Models (LLMs) as a general writing aid. Its role was limited to suggesting minor improvements to grammar, clarity, and LaTeX formatting. All research ideas, algorithmic developments, data analyses, and conclusions 
% were conceived and carried out entirely by the authors. The final content and wording were reviewed 
% and approved by all authors.

\section{Use of Large Language Models (LLMs)}
\label{sup:LLM}
In preparing this manuscript, we used large language models solely as a lightweight writing aid for grammar, wording, and formatting suggestions. The models were \emph{not} used to generate research ideas, design algorithms, write code, run experiments, analyze data, or draft scientific content. All technical claims, methods, and conclusions were conceived, produced, and verified by the authors. Suggested edits from LLMs were manually reviewed and integrated at the authors' discretion. And we accept full responsibility for the accuracy and integrity of the manuscript, including ensuring that no plagiarized or misrepresented content from a LLM is included.

\section{More Qualitative results for integration with CFG}
\label{sup:combination}

\begin{figure}[htp]
  \centering
  \includegraphics[width =\linewidth]{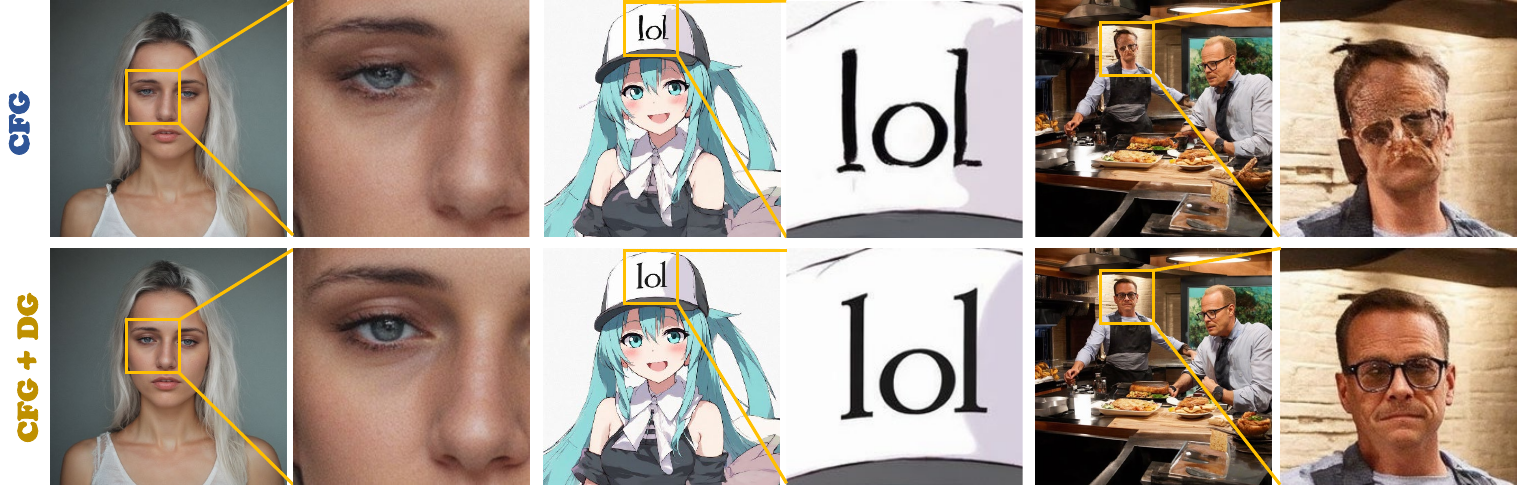}
  \caption{\textbf{Visual results on SD3.}
  % DiTs Only achieve performances of approximately 40 on PCK@0.10 metric in the Spair-71k dataset, while the performance of SD2-1(Unet) is about 60.
  }
  % \vspace{-4mm}
  \label{figure:sd3_sup_v1}
\end{figure}

\begin{figure}[htp]
  \centering
  \includegraphics[width =\linewidth]{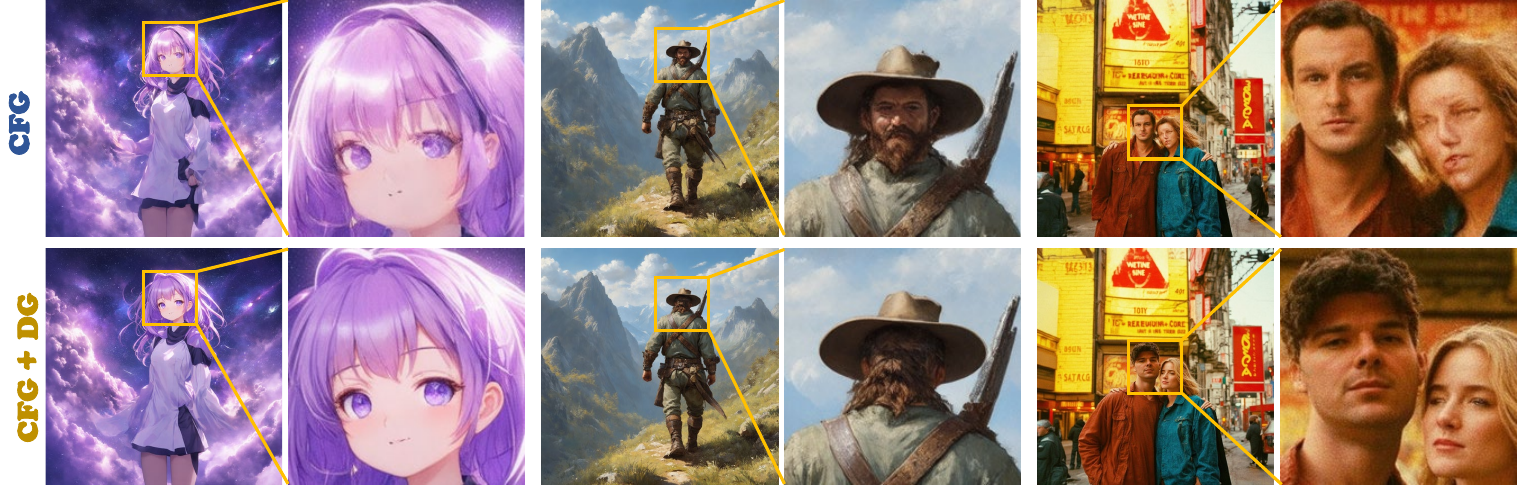}
  \caption{\textbf{Visual results on SD3.}
  % DiTs Only achieve performances of approximately 40 on PCK@0.10 metric in the Spair-71k dataset, while the performance of SD2-1(Unet) is about 60.
  }
  % \vspace{-4mm}
  \label{figure:sd35_sup_v2}
\end{figure}

\begin{figure}[htp]
  \centering
  \includegraphics[width =\linewidth]{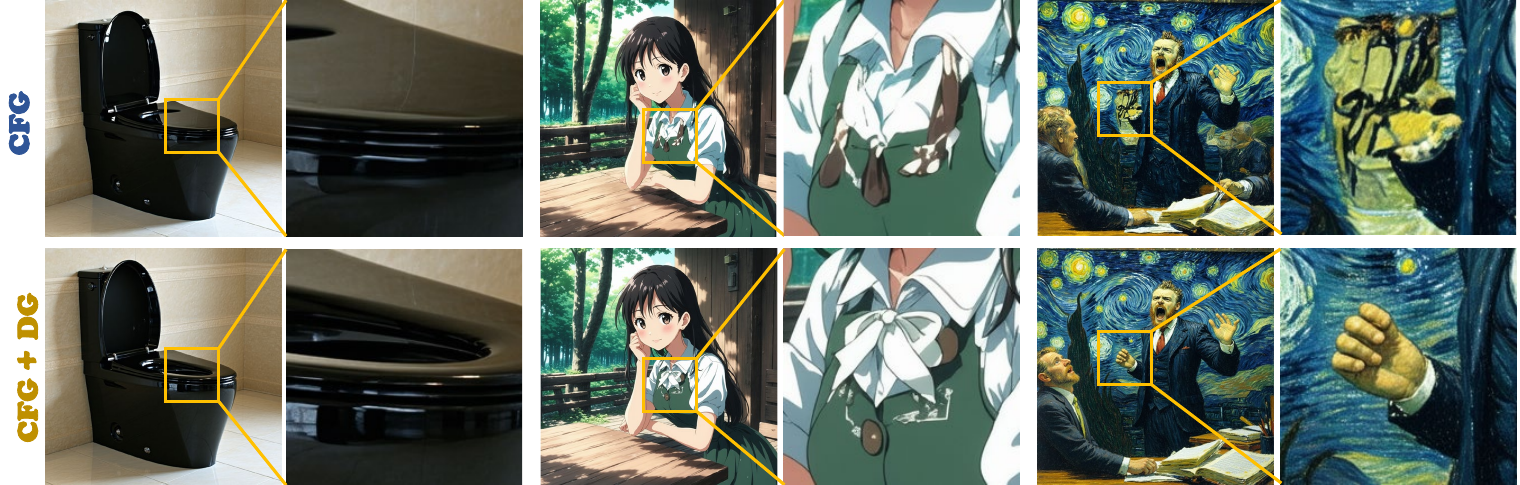}
  \caption{\textbf{Visual results on SD3.5.}
  % DiTs Only achieve performances of approximately 40 on PCK@0.10 metric in the Spair-71k dataset, while the performance of SD2-1(Unet) is about 60.
  }
  % \vspace{-4mm}
  \label{figure:sd35_sup_v1}
\end{figure}
\newpage
\section{More Qualitative results for DG}
\label{sup:qualitative}
\begin{figure}[htp]
  \centering
  \includegraphics[width =\linewidth]{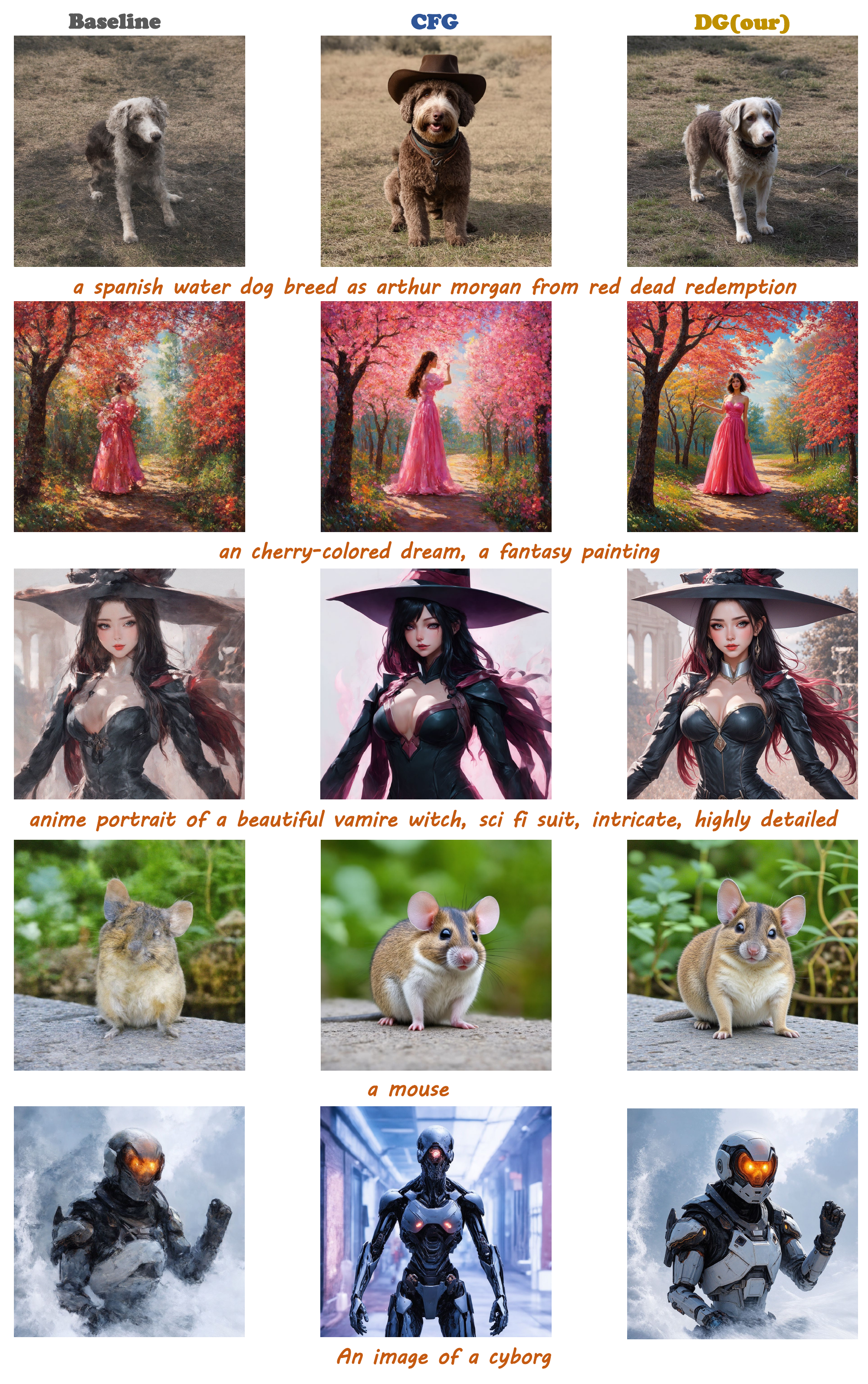}
  \caption{\textbf{Visual results on SD3.} Baseline indicates visual output without CFG.
  % DiTs Only achieve performances of approximately 40 on PCK@0.10 metric in the Spair-71k dataset, while the performance of SD2-1(Unet) is about 60.
  }
  \vspace{-4mm}
  \label{figure:sd3_sup}
\end{figure}

\begin{figure}[htp]
  \centering
  \includegraphics[width =\linewidth]{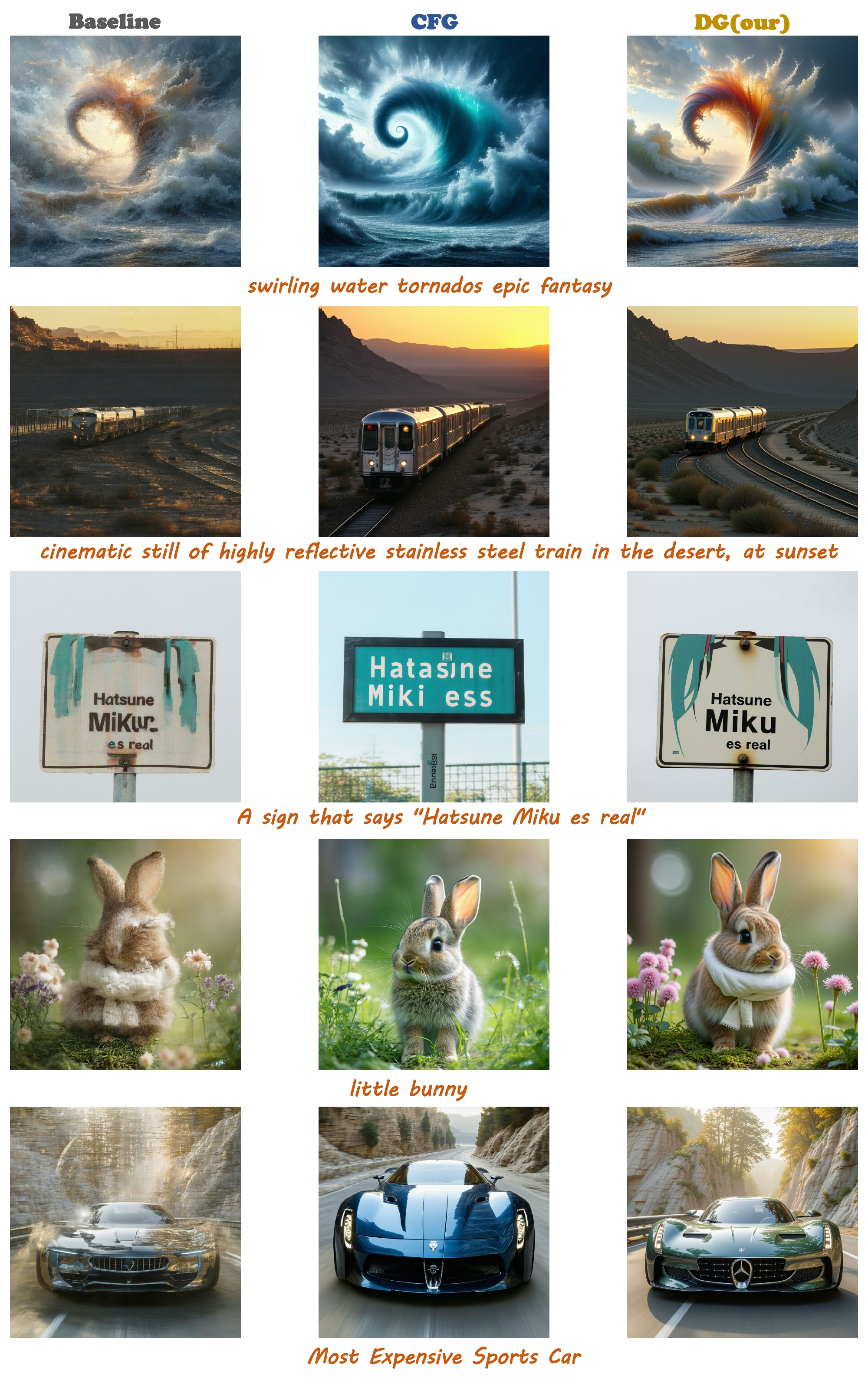}
  \caption{\textbf{Visual results on SD3.5.} Baseline indicates visual output without CFG.
  % DiTs Only achieve performances of approximately 40 on PCK@0.10 metric in the Spair-71k dataset, while the performance of SD2-1(Unet) is about 60.
  }
  \vspace{-4mm}
  \label{figure:sd35_sup}
\end{figure}

\begin{figure}[htp]
  \centering
  \includegraphics[width =\linewidth]{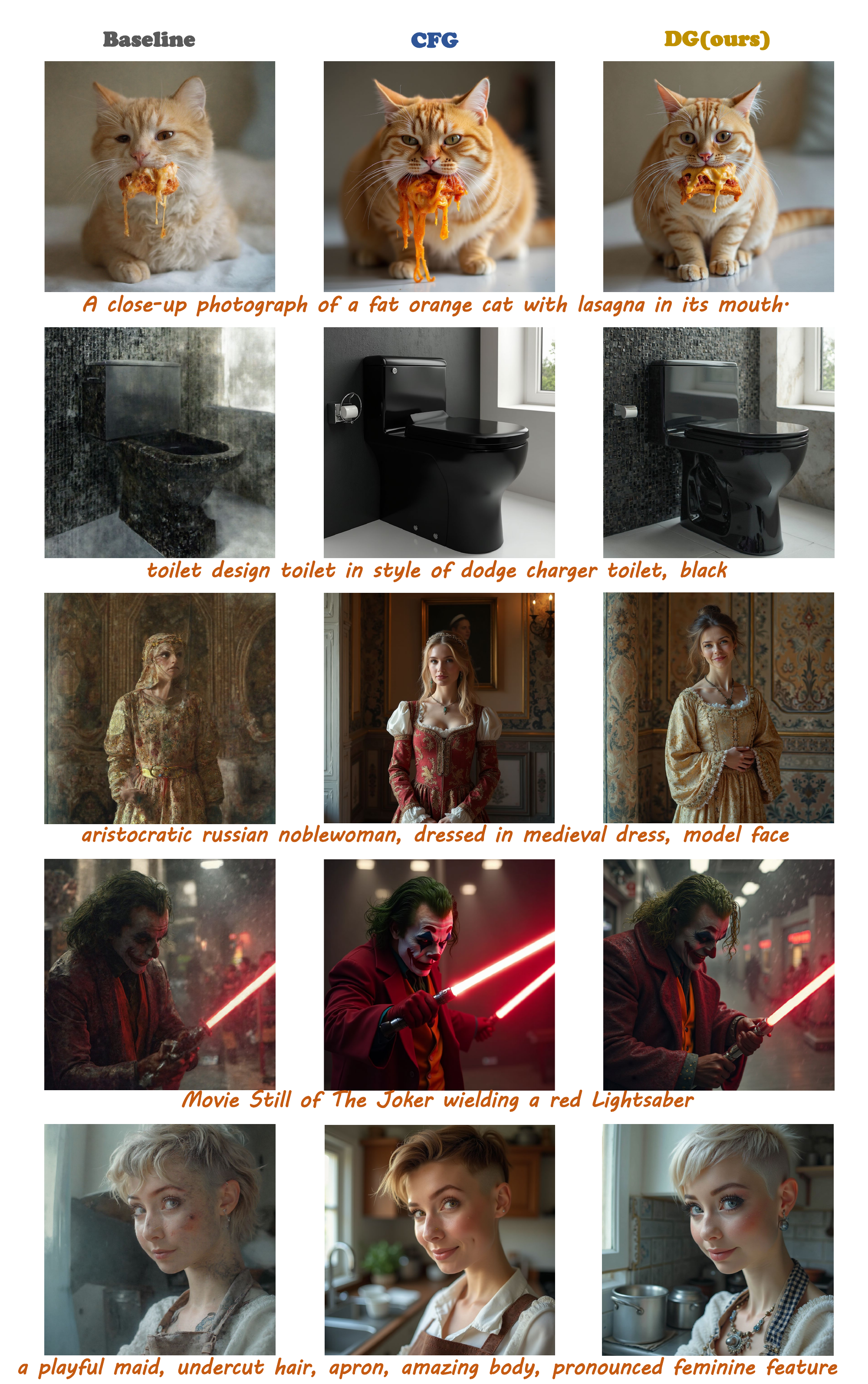}
  \caption{\textbf{Visual results on Flux.} Baseline indicates visual output without CFG.
  % DiTs Only achieve performances of approximately 40 on PCK@0.10 metric in the Spair-71k dataset, while the performance of SD2-1(Unet) is about 60.
  }
  \vspace{-4mm}
  \label{figure:flux_sup}
\end{figure}

\end{document}